\documentclass{article}

\usepackage{PRIMEarxiv}

\usepackage[utf8]{inputenc} % allow utf-8 input
\usepackage[T1]{fontenc}    % use 8-bit T1 fonts
\usepackage{url}            % simple URL typesetting
\usepackage{booktabs}       % professional-quality tables
\usepackage{amsfonts}       % blackboard math symbols
\usepackage{nicefrac}       % compact symbols for 1/2, etc.
\usepackage{microtype}      % microtypography
\usepackage{lipsum}
\usepackage{fancyhdr}       % header
\usepackage{graphicx}       % graphics

\usepackage{graphicx} 
\usepackage{subcaption}
\usepackage{tcolorbox}
\usepackage{tabularray}
\usepackage{array}
\usepackage{makecell}
\usepackage{tabularx}
\usepackage{colortbl}
\usepackage{hhline}   
\usepackage{multirow} 
\usepackage{booktabs}
\usepackage{amsmath}  
\usepackage{amsfonts}   
\usepackage{amssymb} 
\usepackage{mathtools}
\usepackage{bm}% bold math
\usepackage{subfiles}
\usepackage{blindtext}
\usepackage{minitoc}
\usepackage{setspace}
\usepackage{endnotes} 
\usepackage{fancyvrb} 
\usepackage{rotating}

\usepackage{dcolumn}% Align table columns on decimal point
\usepackage[utf8]{inputenc}
\usepackage[T1]{fontenc}
\usepackage{etoolbox}

\usepackage{algorithmic}
\usepackage{float}

\usepackage{xcolor}
\usepackage{tikz}

\usepackage{epstopdf, epsfig}
\usepackage{calc}
\usepackage{microtype}
\usepackage{lscape}
\usepackage{esint,upgreek}
\usepackage{afterpage}
\usepackage{listings}

\usepackage{framed} % Framing content
\usepackage{multicol} % Multiple columns environment
\usepackage{nomencl} % Nomenclature package

\usepackage{algorithm2e}
\RestyleAlgo{ruled}

\usepackage[normalem]{ulem}

\usepackage[numbers]{natbib}

\usepackage[normalem]{ulem}
\usepackage{color}
\usepackage{utils/mylines}
\usepackage{utils/myplots}
\newcommand{\sy}[2]{\mbox{(\kern-.25em\SymbolRGB[solid]{#1}{.8pt}{#2}{5pt}\kern-.25em)}}
\newcommand{\lsy}[3]{\mbox{(\kern-.1em\lineSymbolRGB{#1}{#2}{2pt}{#3}{4pt}\kern-.45em)}}
\newcommand{\lcap}[2]{~\,{\kern-1em\protect\mylcap{#1}{#2}}}

\definecolor{blue}{rgb}{0,0,1}
\definecolor{red}{rgb}{1,0,0}
\definecolor{black}{rgb}{0,0,0}
\definecolor{white}{rgb}{1,1,1}
\definecolor{greyR}{RGB}{50,50,50}
\definecolor{redR}{RGB}{227, 47.3333, 39}
\definecolor{greenR}{RGB}{55, 160.3333, 85}
\definecolor{blueR}{RGB}{55, 135, 192.3333}
\definecolor{yellowR}{rgb}{0.9412, 0.7843, 0.0588}

\definecolor{greyRr}{RGB}{188,188,188}
\definecolor{redRr}{RGB}{239, 182, 182}
\definecolor{blueRr}{RGB}{179, 223, 241}
\definecolor{greenRr}{RGB}{180, 245, 217}

\newcommand{\HYGO}{\textsf{HyGO}\xspace}

\definecolor{myorange}{HTML}{D95319}
\definecolor{myblue}{HTML}{0072BD}
\definecolor{myyellow}{HTML}{EDB120}
\definecolor{mygreen}{HTML}{77AC30}

\definecolor{color_rebut}{RGB}{20, 20, 255}

\definecolor{color_rebut2}{RGB}{50, 205, 100}

\usepackage{hyperref}
\definecolor{pastelblue}{RGB}{90,130,190}
\hypersetup{
    colorlinks=true,
    linkcolor=pastelblue,
    citecolor=pastelblue,
    urlcolor=pastelblue,
    filecolor=pastelblue,
    pdfborder={0 0 0}
}

\graphicspath{{media/}}     % organize your images and other figures under media/ folder

\title{Fast and robust parametric and functional learning with Hybrid Genetic Optimisation (\HYGO)
}

\author{
  Isaac Robledo \\
  Department of Aerospace Engineering \\
  Universidad Carlos III de Madrid \\
  Madrid, Spain \\
  \texttt{isaac.robledo@alumnos.uc3m.es} \\
   \And
  Yiqing Li \\
  Department of Mechanical Engineering \\
  University College London \\
  London, United Kingdom \\
  \texttt{yiqing.li@ucl.ac.uk} \\
  \And
  Guy Y. Cornejo Maceda \\
  Department of Aerospace Engineering \\
  Universidad Carlos III de Madrid \\
  Madrid, Spain \\
  \texttt{gcornejo@ing.uc3m.es} \\
  \And
  Rodrigo Castellanos \\
  Department of Aerospace Engineering \\
  Universidad Carlos III de Madrid \\
  Madrid, Spain \\
  \texttt{rcastell@ing.uc3m.es} \\
}

\begin{document}
\maketitle
\begin{abstract}
The Hybrid Genetic Optimisation framework (\HYGO) is introduced to meet the pressing need for efficient and unified optimisation frameworks that support both parametric and functional learning in complex engineering problems. Evolutionary algorithms are widely employed as derivative-free global optimisation methods but often suffer from slow convergence rates, especially during late-stage learning. 
\HYGO integrates the global exploration capabilities of evolutionary algorithms with accelerated local search for robust solution refinement.
The key enabler is a two-stage strategy that balances exploration and exploitation.
For parametric problems, \HYGO alternates between genetic algorithm and targeted improvement through a degeneracy-proof Dowhill Simplex Method (DSM). 
For function optimisation tasks, \HYGO rotates between genetic programming and DSM.
Validation is performed on (a) parametric optimisation benchmarks, where \HYGO demonstrates faster and more robust convergence than standard genetic algorithms, and (b) function optimisation tasks, including control of a damped Landau oscillator. 
Practical relevance is showcased through aerodynamic drag reduction of an Ahmed body via Reynolds-Averaged Navier-Stokes simulations, achieving consistently interpretable results and reductions exceeding 20\% by controlled jet injection in the back of the body for flow reattachment and separation bubble reduction. Overall, \HYGO emerges as a versatile hybrid optimisation framework suitable for a broad spectrum of engineering and scientific problems involving parametric and functional learning.
\end{abstract}

% keywords can be removed
\keywords{Hybrid optimisation \and  Genetic Algorithms \and Genetic programming \and high-dimensional optimisation  \and Downhill Simplex Method \and Flow control \and Drag reduction \and Ahmed body}

%-------------------------------------------------------------
%----------------- INTRODUCTION ------------------------------
%-------------------------------------------------------------
\section{Introduction} \label{sec:Intro}
Global optimisation problems, particularly those characterised by strong nonlinearity, high dimensionality, and the presence of multiple local minima, remain a central challenge across scientific and engineering disciplines. In engineering, the goal is often to determine the most efficient solution to a well-defined problem, constrained by design, operational, or physical limits. Modern tasks frequently involve high-dimensional design spaces, expensive and time-consuming evaluations, and collinearity between parameters, making them especially prone to entrapment in local optima and difficult to address with gradient-based methods. These factors not only lead to rugged and multimodal optimisation landscapes but also expose algorithms to the curse of dimensionality \citep{Arora1995_Global_opt}, where the volume of the design space grows exponentially with the number of variables, severely hampering convergence and efficiency. Representative examples include aerodynamic shape optimisation of vehicles or aircraft, where even small geometric variations can yield highly non-intuitive aerodynamic responses \citep{li2022ML_shape_opt_review}; structural design under multi-physics constraints, such as in thermomechanical systems or energy absorbers \citep{xie2023thermal_opt}; and control of turbulence through active or passive devices, where system dynamics are governed by nonlinear interactions and time-delays between inputs and outputs \citep{Brunton2015turbulencecontrol}.

% Evolutionary methods and others
To address the challenges posed by high-dimensional, nonlinear, and multimodal global optimisation problems, where purely local (exploitation-driven) methods often fail, various optimisation strategies have been developed to balance broad exploration with focused refinement. These strategies include evolutionary algorithms, such as genetic algorithms and differential evolution, which excel in exploring complex search spaces without reliance on gradient information; gradient-based methods, effective primarily for smooth and convex landscapes; Bayesian optimisers, providing uncertainty estimates; and reinforcement learning approaches that cast optimisation as sequential decision-making with feedback. Each approach presents trade-offs in convergence speed, robustness, scalability, and sensitivity to noise or local minima. Evolutionary algorithms, in particular, are distinguished by their robustness and flexibility in handling black-box, multi-objective, and noisy problems, and are highly adaptable for incorporating real-world constraints through penalty functions or specialised operators, making them highly suited to many engineering optimisation challenges \citep{Rocke2000Genetics}.

% GA
Within the family of evolutionary algorithms, genetic optimisers stand out as a versatile and robust meta-heuristic for exploring complex search spaces. As stochastic methods inspired by natural principles, genetic optimisers operate on populations of candidate solutions that evolve iteratively through selection, crossover, and mutation, effectively balancing exploration of the search space with convergence toward promising optima \citep{Goldberg1989genetics}. Genetic optimisers commonly subdivide into two categories based on solution representation: genetic algorithms (GAs) and genetic programming (GP). GAs operate on fixed-length parametric vectors, making them particularly well-suited for engineering problems defined over structured parameter domains. Their applications range from fluid flow control, such as plasma actuator optimisation for flow reattachment behind a backwards-facing step \citep{Benard2016JetControl}, pulsed jets to minimise aerodynamic drag in a notch-back Ahmed body \citep{he2024active}, upstream actuators for bluff-body wake control \citep{minelli2020upstream}, and heat transfer enhancement in wall-bounded turbulent flows \citep{Castellanos2023ControlGen}, to structural engineering design, including the design of 1D vibration-damped beams \citep{Wu2022lGA_example_structures} and performance-driven architectural elements like building roofs \citep{Turrin2011LGA_structure}. GAs also showcased in medical applications such as disease classification \citep{kumar2020medicine}.

% GP
Conversely, genetic programming evolves variable-length symbolic expressions or programs, which are effective when the optimisation target is a function or decision-making policy. While historically limited by computational cost and complexity, recent algorithmic innovations and parallelisation have expanded GP’s accessibility and impact. GP has gained prominence in adaptive control and optimisation, notably in flow control scenarios such as turbulent jet mixing \citep{zhou2020GP_jet}, wake flow stabilisation \citep{castellanos2022GP}, and drag reduction in complex aerodynamic configurations including Ahmed bodies \citep{li2017dragGP, Li2018am}. Its symbolic formulation capability has also proven valuable in solving ordinary and partial differential equations \citep{Tsoulos2006GP6, Sobester2008GP10}, or in time series forecasting \citep{Wagner2007GP11}. Beyond engineering, GP’s symbolic and interpretable nature has made it useful in finance \citep{WANG2022_credit_porfolio_opt} and medicine \citep{kumar2020medicine}, highlighting its versatility for producing adaptive, interpretable functional strategies.

% Improvements of genetic optimisers and main limitations
Over the years, the widespread adoption of genetic algorithms has driven numerous methodological improvements aimed at enhancing their efficiency and reliability. These include advanced selection mechanisms \citep{Holland1992Mutation}, adaptive crossover techniques \citep{Syswerda1989Crossover, Wright1991GAParametric, Deb1995Crossover2}, robust mutation operators \citep{Deb1996Mutation5, Beyer2001Mutation4, Goldberg1985Mutation3, Syswerda1991Mutation2}, and dynamic parameter tuning strategies designed to adapt search behaviour over time. Despite these advances, both GA and GP share a fundamental limitation: their reliance on stochastic, population-based exploration often results in slow convergence and limited local refinement. Their inherent exploratory nature tends to lead to inefficiency, with the risk of premature convergence to suboptimal solutions.

% Hybrid Genetic Optimisers
To mitigate these issues, recent developments have focused on hybridising genetic algorithms with local search techniques, effectively combining broad global exploration with targeted fine-tuning. Such hybrids aim to accelerate convergence toward near-global optimum, reduce sensitivity to initial conditions, and prevent premature convergence to local minima. For example, hybridising genetic optimisers with gradient-based methods can significantly improve convergence rates, particularly in smooth and differentiable landscapes \citep{Nocedal2006NumOpt}. Alternatively, integrating local search algorithms \citep{Hoos2004LocalSearch2} or stochastic refinement strategies such as Simulated Annealing \citep{adler1993geneticSimulatedAnnealing} enhances robustness by enabling broader and more adaptive exploration of the solution space.

% Introduction to DSM
A particularly successful algorithm for local exploitation is the Downhill Simplex Method (DSM), valued for its simplicity, interpretability, robustness, and effectiveness in gradient-free scenarios. DSM has been widely used in hybrid frameworks with evolutionary algorithms \citep{XU2017DSM_hybrid_evolutionary} and genetic algorithms \citep{yang1998GA_DSM}, for applications including geometric and geoacoustic parameter estimation \citep{Musil1999GA_DSM_Geoacustic}, chiller design optimisation \citep{MAEHARA2013GA_DSM_chiller}, and, more recently, in combination with genetic programming for flow control \citep{cornejo2011JFM_gMLC}. Its sustained popularity stems from its ability to complement global genetic strategies with efficient, gradient-free local optimisation, improving convergence behaviour and solution quality in complex, high-dimensional problems.

% Introduction to our solution
By leveraging the complementary strengths of global and local search paradigms, such as combining genetic optimisation with the DSM, hybrid optimisation frameworks offer a more balanced and effective approach to solving complex, high-dimensional engineering problems. These strategies underpin advanced metaheuristics capable of addressing the increasing demands of modern applications in control, design, and optimisation. Building on these principles and motivated by the need for robust, flexible hybrid genetic optimisers, we propose the Hybrid Genetic Optimiser framework (\HYGO), which extends the hybrid mechanisms introduced by \citet{cornejo2011JFM_gMLC}.
\HYGO is designed as a versatile, extensible framework that integrates global and local search strategies, supporting hybridisation with a broad range of local optimiser algorithms such as BFGS and COBYLA, provided suitable parsing between solution representations and local method inputs is ensured. However, this manuscript focuses specifically on hybridisation with DSM due to its widespread acceptance, prior successful use in hybrid frameworks, and to evaluate our proposed degeneracy-proof adaptation. This key innovation mitigates common degeneracy issues in high-dimensional simplices, such as collapse to lower-dimensional manifolds, by implementing corrective measures that preserve the simplex geometry throughout optimisation, thereby enhancing convergence robustness and efficiency.
\HYGO also incorporates robust uncertainty-handling tailored for experimental applications, including repeated individual evaluations and user-defined outlier exclusion.
Moreover, although \HYGO is an unconstrained algorithm, it features a regeneration mechanism that replaces individuals that do not meet user-defined constraints.
This mechanism preserves population diversity while maintaining feasibility.
The framework also features built-in error recovery capabilities, such as automatic check-pointing and secure fallback strategies, enabling optimisation to resume following evaluation failures. Supporting both parametric encodings and Linear Genetic Programming (LGP) formulations, \HYGO is applicable across a wide variety of problem types. Fully implemented in Python, it offers a modular, customisable platform well-suited for integration into contemporary optimisation pipelines in simulation-based and experimental contexts.

% Scenarios of evaluation
To validate the \HYGO framework, a diverse suite of test cases was selected to highlight the strengths of both GA and GP components, as well as the DSM-based local refinement. Initially, the GA component was assessed on standard analytical benchmark functions, affording controlled evaluation of stochastic convergence speed and solution quality. 
Subsequently, the GP component was employed to stabilise the damped Landau oscillator, a canonical test problem modelling the nonlinear dynamics of von Kármán vortex shedding behind a cylinder \citep{Luchtenburg2009Landau}. This case evaluated the hybrid approach's capability to enhance LGP performance through DSM enrichment in a non-trivial, time-dependent dynamical system. Finally, \HYGO was applied to aerodynamic drag minimisation on an Ahmed body, a widely studied geometry in automotive aerodynamics. This real-world application in a simulation-based environment using RANS simulations demonstrated the framework's practicality in high-dimensional design spaces and enabled direct comparison against the optimisation approach proposed by \citet{Li2022JFM_EGM}, providing a comprehensive context for assessing hybrid methodology effectiveness.

% Structure of paper
The remainder of this manuscript is structured as follows. \hyperref[s:Hybrid_Hygo]{Section~\ref*{s:Hybrid_Hygo}} introduces the proposed hybrid genetic optimisation framework, detailing its architecture, genetic components, and local refinement strategy. \hyperref[s:LGA_analytical_func]{Section~\ref*{s:LGA_analytical_func}} evaluates \HYGO’s performance on analytical benchmark functions, comparing its convergence behaviour and robustness against classical optimisation techniques. In \autoref{s:GP_validation}, the framework is applied to the control of the damped Landau oscillator using LGP, serving as a benchmark for time-dependent control tasks. \hyperref[s:furgo]{Section~\ref*{s:furgo}} presents the drag reduction application on an Ahmed body. Finally, \autoref{s:conclusions} concludes the study and outlines directions for future research.

%--------------------------------------------------------------
%----------------------- HYGO ---------------------------------
%--------------------------------------------------------------
\section{Hybrid Optimisation with \HYGO}\label{s:Hybrid_Hygo}
In this section, we introduce \HYGO, our optimisation framework developed to facilitate the design and deployment of hybrid genetic optimisers. The framework offers a structured environment to combine evolutionary algorithms with a variety of local enrichment techniques. We illustrate the methodology by merging parametric and functional optimisers, specifically GA for parametric optimisation and LGP for functional optimisation, with a variant of the DSM for local search purposes. The following subsections detail \HYGO’s architecture, discuss the critical balance between exploration and exploitation in hybrid optimisation, outline the role of evolutionary algorithms in global search, describe the integration of local refinement techniques, and present the specific implementation of our hybrid methodology within the \HYGO framework.

\subsection{Genetic Optimisation Framework} \label{ss:Opt_framework}
Let $J(\mathbf{f};\theta)$ represent the optimisation objective, or cost function, where $\mathbf{f}\in\mathcal{F}$ is a vector of optimisation variables within the feasible search space $\mathcal{F}$, and $\theta$ represents a set of fixed problem parameters defining the environmental or problem conditions, common to all evaluations. The optimisation problem aims to determine the optimal set of variables $\mathbf{f^*}$ such that:

\begin{equation}
\mathbf{f^*} = \arg \min_{\mathbf{f} \in \mathcal{F}} J(\mathbf{f};\theta)
\end{equation}

Genetic optimisers address this problem by emulating natural evolution through \textit{survival of the fittest} mechanisms. In these algorithms, each specific candidate solution within the search space is represented as an \textit{individual}, and a collection of these individuals forms a \textit{population}. The optimisation process is iterative, where each subsequent population $i+1$ is generated from the previous population $i$ by applying evolutionary operators such as replication, crossover, and mutation. This process aims to iteratively improve the population of candidate solutions over successive generations, ultimately converging towards an optimal or near-optimal solution.

The initial population of $I$ individuals is generated using Monte Carlo Sampling (MCS) or alternative methods such as random or Latin Hypercube Sampling (LHS), to ensure diversity. Individuals are evaluated via the cost function and ranked as $J_i^1 \leq J_i^2 \leq \dots \leq J_i^I$. Population evolution relies on four key operations: elitism, replication, mutation, and crossover. Elitism preserves the top $N_e$ individuals, directly passing them to the next generation. For selecting parents for evolutionary operators, tournament selection is employed: randomly selecting $N_t$ individuals from the population, ranking them by cost, and selecting parents probabilistically according to their ranking and a selection probability $p_s$: The individual with the lowest cost is chosen as a parent with probability $p_s$; if this individual is not selected, the second-best is considered with the same probability, and so on, until a parent is selected. This method promotes fitter individuals while maintaining diversity.

\begin{figure*}
    \centering
    \includegraphics[width=0.99\linewidth]{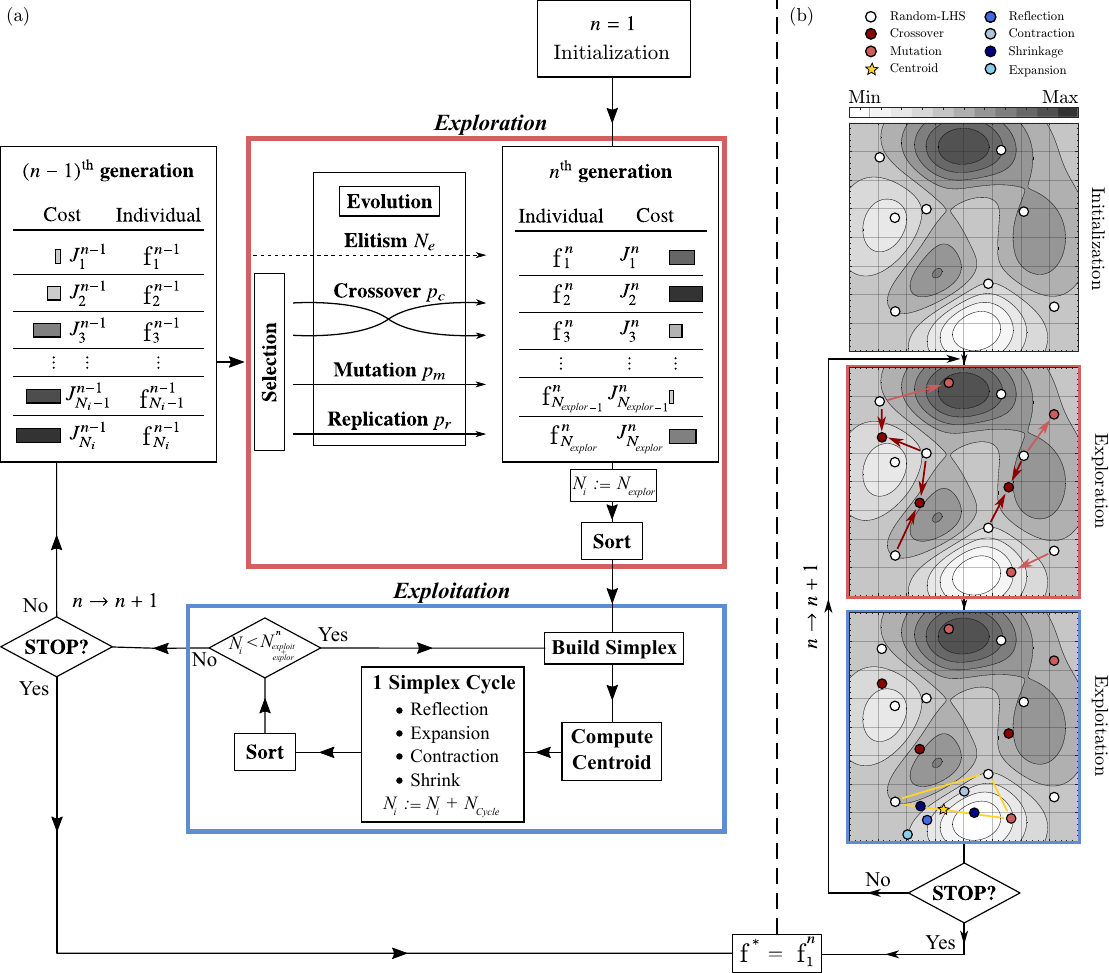}
    \caption{Hybrid Genetic Optimisation algorithm: exploration and exploitation scheme.
    (a) Schematic flowchart, highlighting its two-phase structure. In each generation $n$, exploration is performed by creating individuals $\bm{f}^n_i$ via genetic operations, which include elitism, crossover, mutation, and replication, selected through a tournament process; the resulting solutions are evaluated and sorted according to the cost function $J_i^n$. After exploration, exploitation is performed by applying DSM to refine the best candidates. The horizontal bars visually encode individual performance within the population, with brighter and shorter bars indicating lower (better) costs and darker, longer bars reflecting higher (poorer) solutions.
    (b) Conceptual map illustrating an example of an algorithmic sequence in a two-dimensional landscape, alternating between exploration and local exploitation. Coloured points indicate the origin of each solution: random initialisation, genetic operations (crossover, mutation), and simplex-based moves (reflection, expansion, contraction, shrinkage, centroid). Light-shaded regions correspond to low (optimal) cost areas, while dark regions denote high (suboptimal) cost; the arrows and markers trace how exploration enables the population to sample broadly, while exploitation directs progress toward local minima.}\label{fig:HYGO_Algorithm}
\end{figure*}

Following parent selection, genetic operators are applied probabilistically to generate the next generation of offspring, alongside the elitism individuals. Each selected parent (for replication or mutation) or pair of parents (for crossover) undergoes one of the following operations, based on predefined probabilities $P_r$, $P_m$, and $P_c$ (with $P_r+P_c+P_m=1$). Replication carries over the selected individuals to the next generation, excluding the elite, ensuring that promising solutions remain in the genetic pool. Mutation introduces random changes to a parent's genetic material to effectively explore the search space $\mathcal{B}$. Crossover combines the genetic material of two selected parents to produce one or more new offspring, enabling exploitation of promising solutions. The operation probabilities govern the balance between exploiting individuals from the previous generation and exploring new regions of the search space. The procedure allows iterating through generations until convergence or a maximum number of generations $L$ or evaluations $M=I\times L$ is reached. The general optimisation procedure for genetic optimisers is included in \autoref{fig:HYGO_Algorithm}, bypassing the Exploitation section.

Within this general optimisation framework, the way individuals are represented through their genome primarily influences the algorithm’s behaviour, affecting exploration, exploitation, and the ability to bias the search according to the problem's characteristics. In \HYGO, binary encoding is employed for the GA component mainly for its pragmatic advantages, such as simplicity, computational efficiency, and robustness in implementation. For the functional optimisation component, linear GP is used, leveraging its symbolic and interpretable structure for evolving decision policies. 

% ----------------- GA ---------------------------
\subsubsection{Genetic Algorithm Encoding}\label{sss:GA}

In GAs, the optimisation variables $\mathbf{f}$ are expressed as a vector of $N$ scalar design variables $\mathbf{f} = [f_1, \dots, f_N]^T\in \Omega \subset \mathbb{R}^N$, defined within the domain $\Omega$, where each parameter $f_i$ is defined within a valid interval, $f_i \in [f_{i,\min}, f_{i,\max}]$, $i = 1, \dots, N$, leading to a rectangular parametric domain
\begin{equation}
\Omega = [f_{1,\min},f_{1,\max}] \times \dots \times [f_{N,\min}, f_{N,\max}].
\end{equation}

GA typically employ chromosomes encoding sets of fixed parameters, most commonly using binary encoding due to its simplicity, broad applicability, and compatibility with standard genetic operators such as crossover and mutation, leading to its implementation in \HYGO. Furthermore, binary encoding benefits from efficient memory usage and straightforward implementation, making it particularly suitable for parametric optimisation tasks where candidate solutions are naturally represented as vectors of design variables. While alternative encoding schemes, such as real-valued or Gray code encodings, show potential to enhance precision and improve convergence speed, these alternatives typically require increased complexity in operator design and genotype-phenotype mapping, often without significant benefits in practical scenarios. For this reason, \HYGO adopts binary encoding pragmatically, balancing robust performance with computational efficiency and ease of implementation.

\begin{figure*}
    \centering
    \includegraphics[width=\linewidth]{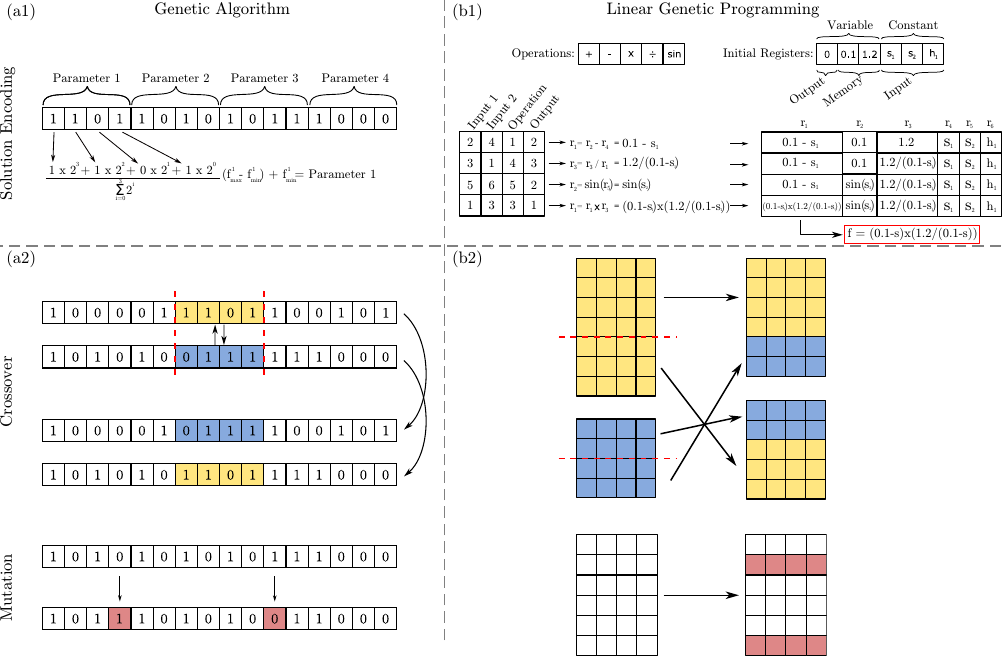}
    \caption{Solution encoding and genetic operations for Genetic Algorithm (GA) and Linear Genetic Programming (LGP). (a1) Binary-encoded chromosomes for GA represent sets of parameters. (b1) Encoding of LGP individuals as instruction matrices acting on program registers (variables, constants, memory, and inputs).
    (a2, b2) Examples of genetic operations: (a2) In GA, crossover swaps chromosome segments between parents (highlighted), and mutation randomly flips selected bits. (b2) In LGP, crossover exchanges blocks of instructions between programs, and mutation replaces random instructions, enabling structural diversity in candidate solutions.
    }\label{fig:Genetic_operations}
\end{figure*}

In binary-encoded GAs as employed in \HYGO, each of the optimisation variables $f_i$ is represented by a string of $n$ binary digits (bits), yielding a resolution of $2^n$ distinct values over the interval $[f_{i,\min}, f_{i,\max}]$. Typically, these values are distributed uniformly, and decoding the binary string yields a corresponding real-valued approximation of the parameter. This encoding enables the straightforward application of genetic operators. The crossover operator partitions two parent chromosomes into $N_S$ segments and interchanges them sequentially, or randomly, to generate one or two offspring. Meanwhile, the mutation operator introduces variation by flipping individual bits in a parent chromosome with a probability $p_m$, producing a new individual and promoting exploration of the search space. A visual summary of the encoding and the genetic operations is provided in \autoref{fig:Genetic_operations} for further clarity.

\subsubsection{Linear Genetic Programming}\label{sss:LGP}
In Genetic Programming, the design variable $\mathbf{f}$ is expressed as an analytical function that depends on a set of numerical constants, external sensor inputs $\mathbf{s}$, and prescribed time-dependent functions $\mathbf{h}$, $\mathbf{f} = \mathbf{f}(\mathbf{h},\mathbf{s})$. Unlike GAs, where individuals are fixed-length vectors, GP encodes individuals as computer programs, providing flexibility for evolving dynamic decision rules or control strategies \cite{Banzhaf1993GP_intro}. Early implementations of GP employed tree-based representations, in which output functions were encoded as hierarchical structures of operations (such as $+,\;-,\;\div,\;\times,\; \max,\; \sin$) and numerical constants. However, these tree-based models often suffered from uncontrolled growth and structural complexity, leading to the development of alternative representations.

A widely used and interpretable variant is \textit{linear} GP, in which individuals are encoded as instruction matrices operating on a shared set of registers. This format allows for compact, symbolic representations of functional laws that evolve over time or in response to sensor inputs, making LGP especially well-suited to problems such as real-time control. As illustrated in \autoref{fig:Genetic_operations} (b1), and following the methodology introduced by \citet{Brameier2007LGP}, each chromosome is expressed as an $N_{it}\times 4$ instruction matrix, with each row specifying a single instruction that combines elements from a set of registers, which include sensor readings $s_i$, time-dependent functions $h_i$, intermediate variables, and outputs. Each instruction comprises four fields (columns): the first two columns specify the input arguments (if the operation is unary, the second input is ignored), the third column indicates the operator (e.g., addition, multiplication, sine), and the fourth column specifies the destination register where the result is stored.

All individuals share initial input registers containing: $N_{\text{out}}$ output registers, which hold the resulting $\mathbf{f}$ vector for each individual; $N_{\text{mem}}$ memory registers, used to store intermediate results and enable complex instruction interactions; and $N_{\text{in}}$ input registers, which provide access to all sensor readings and time-dependent functions used in the optimisation. Through sequential execution, the instruction set produces a program that dynamically maps system states to output functions $\mathbf{f}$. Registers can be either variable, allowing instructions to modify their contents, or constant, making their values accessible to every instruction without modification. Typically, sensors, time-dependent functions, and predefined constants are stored in constant registers, while the output register must always be variable. The sequential nature of instruction execution is what defines the method as \textit{linear}. An example illustrating how a functional output $\mathbf{f}$ is constructed using this matrix-based representation is shown in \autoref{fig:Genetic_operations}.

The evolutionary process in LGP follows principles similar to those used in binary-encoded GAs, while preserving instruction-level coherence. The crossover operator exchanges subsets of instructions between two parent individuals, allowing for structural recombination without disrupting the logic of instruction sequences. Unlike fixed-length binary encodings, LGP permits individuals to have varying chromosome lengths (up to a maximum of $N_{it}$), introducing additional diversity into the population. The mutation operator modifies instructions with a probability $P_m$, replacing them with newly generated random instructions, which further enhances exploration of the solution space.

%----------------------- Exploration vs Exploitation ---------------------
\subsection{Exploration vs Exploitation}
Genetic optimisation is commonly classified as a global optimisation technique due to its inherent ability to combine \textit{exploration}, the broad search of the solution space, with \textit{exploitation}, the local refinement of promising solutions. This dual capability makes evolutionary algorithms particularly attractive for solving complex optimisation problems, especially where traditional gradient-based or local methods struggle to escape local minima or require smoothness and convexity in the objective function. Various practical scenarios highlight the necessity of balancing these two strategies. Large plateaus in the cost landscape require strong exploration, as the absence of gradient information provides little guidance for local descent. Conversely, narrow valleys or funnel-shaped regions benefit from exploitative steps that efficiently guide the search toward optima. Additionally, shallow gradients may allow convergence but at prohibitive computational cost, underscoring the need for exploration to ``jump'' toward steeper descent regions and accelerate optimisation.

The genotype-to-phenotype mapping in genetic optimisation amplifies sensitivity to small genetic changes, promoting effective exploration. Minor modifications in encoded individuals, such as bit flips in chromosomes or instruction changes in programs, can yield dramatically different solutions. This sensitivity is especially pronounced in control law design, where solutions close in actuation space may correspond to distant genetic representations. Consequently, GA and GP inherently favour exploration over exploitation, excelling at generating diverse candidate solutions but often requiring auxiliary mechanisms for effective local refinement.

Traditionally, balancing exploration and exploitation in genetic optimisation relies primarily on tuning the probabilities of replication, mutation, and crossover. Mutation introduces random variations, enhancing exploration, with the mutation probability $P_m$ modulating exploratory intensity. Conversely, crossover combines genetic material from parents, promoting exploitation by steering the population toward local optima. Replication’s probability $P_r$ influences the allocation of individuals to previously explored regions, typically reducing exploration without directly enhancing exploitation. The mutation rate within the genome controls offspring deviation from parents, enabling jumps across plateaus or local minima; excessive mutation, however, risks random search behaviour that can undermine population knowledge. Tournament selection further affects this balance; larger tournament sizes or lower selection probabilities increase exploration by allowing less-fit individuals reproductive chances, while smaller sizes intensify exploitation by favouring fitter individuals.

Recently, the focus has shifted from solely tuning hyperparameters toward hybrid algorithms combining genetic optimisers with complementary techniques to enhance performance. While genetic optimisers proficiently explore vast, complex search spaces, their capacity for efficient local exploitation is often constrained by the suboptimal nature of standard crossover operations. Though various crossover improvements exist \citep{Syswerda1989Crossover, Deb1995Crossover2}, these alone rarely guarantee rapid, precise convergence. To overcome this, hybridisation with local search methods has become the predominant strategy. Integrating genetic global search with precise, speedy local refinement enables algorithms to comprehensively explore solution spaces while effectively exploiting local optima. Such hybrid frameworks demonstrate superior convergence rates and solution quality across diverse complex optimisation problems, solidifying their status as powerful tools in engineering and scientific computation.

%------------------------ HyGO hybridisation ----------------------------
\begin{algorithm}
    \caption{\HYGO pseudo-code}
    \label{alg:hygo}
    \begin{algorithmic}[1]
    \item[] \hfill\textit{--- Initialisation ---}\hfill 
    %\STATE $g = 1$
    \STATE Generate $N_{explor}$ individuals randomly or with LHS $\rightarrow \mathbf{f}^1_r , \;\; r=1,2,\ldots,N_{explor}$
    \STATE Evaluate the fitness of each individual $\rightarrow J(\mathbf{f}^1_r;\theta) , \;\; r=1,2,\ldots,N_{explor}$.
    \STATE Sort the population according to cost $\rightarrow \{J^1_1<J^1_2<\ldots<J^1_{N_{explor}}\}$.
    \item[] \hfill\textit{--- Exploitative stage ---}\hfill 
    \STATE $N_{ind}^1 = N_{explor}$
    \WHILE{$N_{ind}^1<(N_{explor}+N_{exploit})$}{
        \STATE Generate individual with the local search method $\rightarrow \mathbf{f}^1_i$
        \STATE Evaluate the fitness of offspring $\rightarrow J(\mathbf{f}^1_i;\theta)$
        \STATE $N_{ind}^1 := N_{ind}^1 + N_{cycle}$
        }
    \ENDWHILE
    \STATE Sort the population according to cost $\rightarrow \{J^1_1<J^1_2<\ldots<J^1_{N_{explor} + N_{exploit}}\}$.
    \STATE $g = 2$
    %\STATE Build initial simplex from the available individuals
    %\STATE Perform Downhill Simplex operations until $N_t$ individuals are generated
    \item[]
    \WHILE{not convergence \textbf{and} $g<N_g$}{
        \item[] \hfill\textit{--- Explorative stage ---}\hfill 
        \FOR{$i=1$ to $N_{explor}$}{
            \STATE Perform tournament selection process to select parents
            \STATE Generate an individual by elitism/replication/crossover/mutation $\rightarrow \mathbf{f}^g_i$
        }
        \ENDFOR
    
        \STATE Evaluate the fitness of each individual in the population $\rightarrow J(\mathbf{f}^g_r;\theta) , \;\; r=1,2,\ldots,N_{explor}$.
        \STATE Sort the population according to cost $\rightarrow \{J^g_1<J^g_2<\ldots<J^g_{N_{explor}}\}$
        \item[] \hfill\textit{--- Exploitative stage ---}\hfill 
        \STATE $N_{ind}^g = N_{explor}$
        \WHILE{$N_{ind}^g<(N_{explor}+N_{exploit})$}{
            \STATE Generate individual with local search $\rightarrow \mathbf{f}^g_i$
            \STATE Evaluate the fitness of offspring $\rightarrow J(\mathbf{f}^g_i;\theta)$
            \STATE $N_{ind}^g := N_{ind}^g + N_{cycle}$
            }
        \ENDWHILE
        \STATE Sort the population according to cost $\rightarrow \{J^g_1<J^g_2<\ldots<J^g_{N_{explor} + N_{exploit}}\}$.
        \STATE $g = g+1$
        }
    \ENDWHILE
    \item[]
    \STATE Return the best solution $\rightarrow \mathbf{f^*} = \arg \min_{\mathbf{f}} J(\mathbf{f};\theta)$.
    \end{algorithmic}
\end{algorithm}
\subsection{Hybridising Genetic Optimisers with local search algorithms}
Building on the foundational concepts of exploration and exploitation and motivated by the limitations of genetic optimisers in efficiently refining local optima, this subsection introduces the hybrid optimisation methodology proposed in this work. Our approach synergistically combines the broad global search capabilities of binary-encoded GAs and LGP with the precise local refinement offered by local search techniques. The hybridisation is designed to retain the extensive exploratory strength of evolutionary algorithms while significantly enhancing local exploitation, leading to improved convergence speed and solution quality.

Each optimisation generation is divided into two distinct phases. The \textit{explorative} phase uses genetic operations to generate $N_r$ new individuals from the current population, while the \textit{exploitative} phase applies local search algorithms to produce $N_{exploit}$ refined individuals. This structured alternation effectively balances exploration and exploitation within every generation. A detailed description of the algorithm is provided in Algorithm \ref{alg:hygo}, accompanied by a simplified flowchart in \autoref{fig:HYGO_Algorithm}.

A critical challenge in this iterative hybrid process is preserving genetic diversity throughout evolutionary cycles, which is essential for maintaining robust exploration. Genetic operations (as described in \autoref{ss:Opt_framework}) are inherently stochastic, involving random crossover points and mutation flips. As optimisation progresses and diversity diminishes, the probability of generating individuals identical to those already in the population increases, undermining search effectiveness. Such duplicate individuals waste computational resources and reduce the genetic variability vital for efficient search performance. To address this, newly generated offspring resulting from crossover or mutation undergo a validity check against the current population. If duplicates are detected, these individuals are discarded and regenerated, with regeneration attempts capped to prevent infinite loops.

This validity-checking mechanism extends naturally to enforce \textit{soft constraints}, an enduring challenge in genetic optimisation \citep{michalewicz1991handling_Constraints_GA,homaifar1994constrainedGA,Tarkowski2022ConstrainedGA}. The term ``soft'' reflects that strict constraint satisfaction cannot always be guaranteed due to limited regeneration attempts; when constraints cannot be met, individuals may receive penalising extreme fitness costs or be evaluated regardless. However, such situations typically arise only in late generations once genetic diversity is low and convergence is near completion, minimising their practical impact on optimisation robustness.

%---------------- DSM --------------------
\subsubsection*{Downhill Simplex Method: A Robust Gradient-Free Optimisation Technique}
The Downhill Simplex Method, originally proposed by \citet{Nelder1965DSM}, is a widely used optimisation algorithm notable for its simplicity and robustness. Unlike gradient-based methods, DSM does not require explicit gradient information, making it especially well-suited to optimisation problems where derivatives are unavailable or costly to compute. This characteristic positions DSM as an effective local search algorithm within the \HYGO framework, primarily to refine candidate solutions identified by global search heuristics.

\begin{figure}[t]
    \centerline{\includegraphics[width=0.5\linewidth]{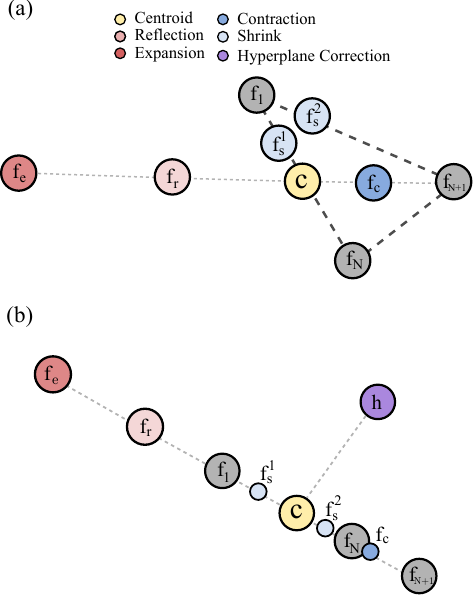}}
    \caption{Hyperplane degeneracy into a 1-dimensional hyperplane in a 2D parametric domain. Image (a) displays all the operations and the topology under normal conditions. Image (b) shows a degenerate scenario where the Simplex collapsed into a line and lost the ability to move in the normal direction.}\label{fig:hyperplane}
\end{figure}

DSM operates by constructing a simplex, a geometric figure composed of $N+1$ vertices in an $N$-dimensional parameter space, starting with an initial simplex spanning the search domain $\Omega$. Each iteration seeks to improve the simplex by replacing the vertex with the highest cost, $\mathbf{f}_{N+1}$, with a new, improved vertex $\mathbf{f}_{N+2}$. The iteration proceeds through the following steps and is represented geometrically in \autoref{fig:hyperplane}(a):
\begin{enumerate}
    \item \textbf{Ordering:} Vertices are sorted by their cost $J_m = J(\mathbf{f}_m):\; J_1 \leq J_2 \leq \dots \leq J_{N+1}$.
    \item \textbf{Centroid Calculation:} Compute the centroid $\mathbf{c}$ of the simplex, excluding the worst vertex $\mathbf{f}_{N+1}$:
    \begin{equation}
        \mathbf{c} = \frac{1}{N} \sum_{m=1}^{N} \mathbf{f}_m.
    \end{equation}
    \item \textbf{Reflection:} The worst vertex $\mathbf{f}_{N+1}$ is reflected about the centroid $\mathbf{c}$ to generate a new vertex $\mathbf{f}_r$:
    \begin{equation}
        \mathbf{f}_r = \mathbf{c} + (\mathbf{c} - \mathbf{f}_{N+1}).
    \end{equation}
    If the cost of the reflected vertex $J_r = J(\mathbf{f}_r)$ lies between $J_1$ and $J_N$ ($J_1 \leq J_r \leq J_N$), replace $\mathbf{f}_{N+1}$ with $\mathbf{f}_r$ and proceed to the next iteration.
    \item \textbf{Expansion:} If reflection yields a new best point $J_r < J_1$, the simplex is expanded further in this direction:
    \begin{equation}
        \mathbf{f}_e = \mathbf{c} + 2 (\mathbf{c} - \mathbf{f}_{N+1}).
    \end{equation}
    The vertex with the lower cost between $\mathbf{f}_r$ and $\mathbf{f}_e$ replaces $\mathbf{f}_{N+1}$, and a new iteration begins.
    \item \textbf{Contraction:} If reflection fails to improve the cost $J_r \geq J_N$, the simplex contracts by moving the worst vertex halfway towards the centroid:
    \begin{equation}
        \mathbf{f}_c = \mathbf{c} + \frac{1}{2} (\mathbf{f}_{N+1} - \mathbf{c}).
    \end{equation}
    If $\mathbf{f}_c$ improves the cost, it replaces $\mathbf{f}_{N+1}$, and the next iteration starts.
    \item \textbf{Shrinkage:} If neither reflection nor contraction reduces cost, the simplex is shrunk by moving each vertex halfway towards the best vertex $\mathbf{f}_1$:
    \begin{equation}
        \mathbf{f}_m \rightarrow \mathbf{f}_1 + \frac{1}{2} (\mathbf{f}_m - \mathbf{f}_1), \quad m = 2, \dots, N+1.
    \end{equation}
    Shrinking is a last-resort operation, as it requires $N$ additional function evaluations, typically applied when the simplex has degenerated to regain local gradients within a more confined parameter region.
\end{enumerate}

The DSM implementation in \HYGO corresponds to this classical version. While alternative variants exist, the classical DSM is directly applicable to parametric optimisation problems by mapping solutions to their nearest binary-encoded points in $\Omega$. For matrix-encoded LGP individuals, a more sophisticated Subplex algorithm \citep{Rowan1990subplex} is employed, as used in prior work \citep{cornejo2011JFM_gMLC}, though its details are beyond this manuscript’s scope. Eventually, during the exploitation phase, \HYGO executes a series of DSM iterations until generating $N_{exploit}$ individuals. However, the shrink operation may produce more candidates than the prescribed $N_{exploit}$, and truncating would disrupt the intended search dynamics and reduce algorithmic flexibility, so $N_{exploit}$ can naturally vary. Moreover, the initialisation of the simplex is key, directly affecting the effectiveness of DSM. Rather than starting with a canonical simplex with one vertex at the centre of the domain and $N$ vertices distributed along each coordinate axis, \HYGO initiates each simplex from the $N+1$ fittest individuals in the current population. 

The hybridisation philosophy process is common in both the parametric (GA) and functional (LGP) implementations, differing in the local search algorithm; GA is combined with the Downhill Simplex algorithm and LGP with the Downhill Subplex adaptation. The functional version builds upon the implementation of gMLC~\citep{cornejo2011JFM_gMLC}; nonetheless, it differs in two key aspects. First, gMLC abandons the concept of population and employs genetic operators and downhill simplex steps to generate new individuals from the pool of all evaluated individuals, where a fraction of the individuals are generated with genetic operators and the remaining individuals within the population with Dowhill Subplex steps.
\HYGO reintroduces a population that evolves through generations.
Second, there is the regeneration process of the Subplex individuals. In the case of gMLC, the Subplex-generated analytical function is maintained even if the regenerated instruction matrix does not perfectly represent it, whereas \HYGO, prior to the individual's evaluation, performs the regeneration and substitutes the Subplex function with the one represented by the regenerated instruction matrix. Despite the validity of both approaches, the strategy followed in \HYGO ensures representativeness.

\subsubsection*{Degeneracy correction mechanism}

For optimisation problems defined in a parametric space of dimension $N$, the simplex can degenerate into a hyperplane of dimensionality $n<N$.
For instance, for a two-dimensional search space, a triangle may degenerate into a line, as illustrated in \autoref{fig:hyperplane}(b). In such cases, the DSM operations cannot recover the missing directions of movement, resulting in inefficient search behaviour, particularly when the gradient path is nonlinear, as in spiral-shaped functions. To detect degeneracy, \HYGO computes the coefficient of determination $R^2$ of the last $N_f$ simplex-generated individuals relative to a hyperplane. If $R^2$ exceeds a user-defined threshold, indicating near-planarity, a corrective step generates a new individual orthogonal to the hyperplane, placed at a distance determined by the grid resolution in the parametric space. This correction restarts the DSM cycle, rebuilding the simplex with the $N+1$ best individuals to accommodate the new candidate. While this degeneracy correction may introduce extra individuals when the path is strongly directional, modestly increasing computational cost, it substantially enhances search robustness in complex landscapes. Another approach for tackling the DSM degeneracy is by maximising the simplex volume, recently proposed by \citet{wang2025rdsmrobustdownhill}. The authors propose a correction of the degeneracy by solving a nonlinear maximisation problem under constraints. The degeneracy correction we propose in this paper constitutes a practical alternative that is expected to be more efficient in high-dimensional problems in terms of computational load.

Furthermore, an operational issue was detected during the execution of the Downhill Simplex Method or Subplex within the hybridisation loop: the local search procedure occasionally entered a state where no standard simplex operation (reflection, expansion, contraction, shrink, or even correction) resulted in a lower objective function value, thereby incurring no change to the simplex vertices. In standard, standalone DSM implementations, this condition typically triggers the convergence criterion, terminating the optimisation; however, since DSM serves as a local search enhancer and not the primary global optimiser in \HYGO, immediate termination is suboptimal as it wastes the allocated local search budget. To mitigate this stagnation, a ``random correction'' operation was introduced. If no change occurs in the simplex after a predefined number of attempts, a randomly sampled point in the vicinity of the simplex centroid is evaluated. This heuristic introduces a controlled local exploration step, thereby allowing the procedure to continue searching for local improvements without premature convergence.

Both the degeneracy-correction and random correction heuristics are independently evaluated and implemented if their respective trigger conditions are met. During execution, the hyperplane-correction mechanism is assigned priority in the sequence for evaluating and introducing new trial points to the simplex, ensuring that geometric validity is addressed before attempting to recover from stagnation.

It must be highlighted that, due to the analytical nature of the Downhill-Subplex algorithm, the standard hyperplane-degeneracy correction cannot be directly applied to its internal state. Consequently, while the random correction mechanism is fully implemented to prevent stagnation, the more sophisticated hyperplane-based degeneracy correction is reserved for the general DSM implementation and is not functional when using the Subplex method.
%
%---------- Analytical Functions --------
\section{Benchmarking Parametric Optimisation on Complex Analytical Functions} \label{s:LGA_analytical_func}
A suite of well-established analytical benchmark functions, known for their challenging optimisation landscapes, served as the baseline for evaluating the convergence performance of the proposed hybrid GA-DSM optimiser. Optimisations were conducted both with and without the exploitation phase, allowing assessment of the local refinement’s impact. To situate performance in a modern context, the approach was compared against the Covariance Matrix Adaptation Evolution Strategy (CMA-ES), a state-of-the-art evolutionary optimiser which utilises multivariate Gaussian distributions to generate populations and accelerates convergence by exploiting the evolution path \citep{Hansen2001CMA, Hansen2003CMA-ES}; Differential Evolution (DE), a robust and widely-used evolutionary algorithm known for its efficient global search capability based on vector differences \citep{storn1997differential, price2005differential}; Particle Swarm Optimisation (PSO), a classic swarm intelligence technique that models social behaviour to guide particles toward promising regions of the search space \citep{eberhart2001swarm, poli2007analysis}; and L-SHADE, an adaptive differential evolution algorithm characterised by linear population decrease and history-based parameter adaptation that significantly boost performance of traditional DE \citep{Tanabe2014lshade}. The configuration parameters used for \HYGO and GA optimisations are summarised in \autoref{Tab:Parameters}.

%----------------- Preliminary assesment: Rosenbrok ---------
\subsection{Preliminary benchmark: Rosenbrock 2D} \label{ss:LGA_analytical_preliminary}
\begin{figure*}
    \centering
    \includegraphics[width=\linewidth]{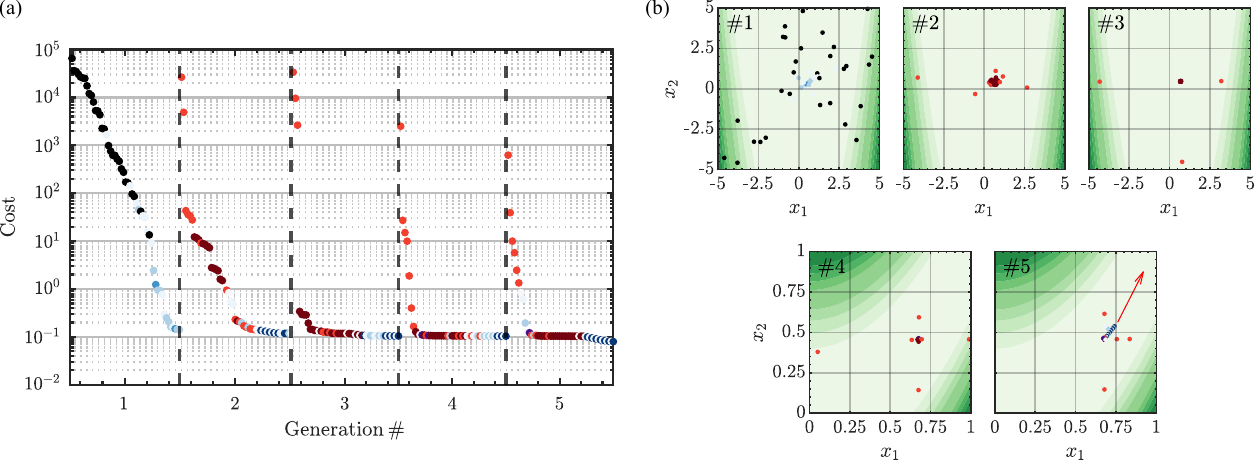}
    \caption{Example of \HYGO performance on the optimisation of the Rosenbrock function in 2D: (a) cost evolution of the individuals through the generations. (b) Individuals' distribution in the parametric space in each of the five generations ($\#1-5$). The arrow in generation \#5 shows the tendency of the DSM solution towards the global minimum of the Rosenbrock function at (1,1). Individuals are coloured by the type of operation that created each individual (see \autoref{fig:HYGO_Algorithm}, black for random initialisation, red tones for the different genetic operations, and blue tones for the DSM operations).} \label{fig:J_evo_preliminary_opt_LGA}
\end{figure*}

To illustrate the convergence improvements obtained by incorporating DSM local search, an optimisation of the 2D Rosenbrock function was performed. This experiment comprised five generations of 50 individuals each, with 30 generated via the explorative phase and 20 produced by the DSM-driven exploitative phase. The initial population of 30 individuals was sampled uniformly at random within the bounds $x_i\in\left[-5,5\right]$, depicted as black dots in the first generation in \autoref{fig:J_evo_preliminary_opt_LGA}.

In \autoref{fig:J_evo_preliminary_opt_LGA}, coloured dots represent the origin of each individual: red-toned dots correspond to explorative GA individuals, and blue-toned dots indicate exploitative DSM-refined individuals. The figure graphically demonstrates the acceleration in convergence enabled by the exploitation phase; exploitative individuals consistently occupy the lower cost region of the population distribution in every generation (panel (a)). The first generation particularly highlights these gains, as large gradients distant from the Rosenbrock valley allow the DSM to significantly reduce cost. As optimisation advances, improvements slow due to shallow gradients near the Rosenbrock valley minimum at $x_i^*=\left[1,1\right]$ (marked by the red arrow in generation \#5, panel (b)). The diminishing gradient magnitudes result in progressively smaller simplex steps, constraining the effectiveness of local refinement in late generations. The distribution plots in panel (b) additionally reveal how exploration and exploitation interplay spatially: explorative offspring are initially more dispersed, while exploitative individuals cluster tightly in promising regions, guiding convergence.

These results visually highlight the complementary strengths of the explorative and exploitative phases in the hybrid optimiser. Nonetheless, performance is subject to enhancement by hyperparameter tuning. For instance, initialising the population via Latin Hypercube Sampling would prevent premature convergence toward a local minimum far from the global optimum, mitigating slow progression down the Rosenbrock valley. Additionally, the current mutation operator employs a conservative \textit{at least one} rule, which sets the bit mutation probability such that most mutated individuals differ by a single bit, altering only one parameter. This restriction limits mutation-induced exploration to vertical or horizontal movements in parameter space (see generation \#4 in \autoref{fig:J_evo_preliminary_opt_LGA}(b)), potentially constraining robustness. Increasing mutation rates could enable diagonal moves and better complement the local search phase.

\begin{table}[t]
    \centering
      \caption{Algorithm configuration parameters for the analytical functions benchmark.} \label{Tab:Parameters}
    \begin{tabular}{@{}cccl@{}}
    \toprule
    \multicolumn{1}{l}{} & Parameter & Value & Description \\ \midrule
    \multirow{7}{*}{\begin{tabular}[c]{@{}c@{}}\HYGO \\ and \\ GA\end{tabular}} & $D$ & 5/25 & Dimension \\
     & $N_b$ & 12 & Bits for each parameter \\
     & $N_G$ & 50 & Total number of generations \\
     & $N_T$ & 7/100 & Tournament size \\
     & $P_c$ & 0.55 & Crossover probability \\
     & $P_m$ & 0.45 & Mutation probability \\
     & $P_r$ & 0 & Replication probability \\ \midrule
    \multirow{3}{*}{\HYGO} & $N_{\rm MC}$ & 70 & Monte Carlo initialisation \\
     & $N_{explor}$ & 70 & Population size (exploration) \\
     & $N_{exploit}$ & 30 & Simplex size (exploitation) \\ \midrule
    \multirow{2}{*}{GA} & $N_{\rm MC}$ & 100 & Monte Carlo initialisation \\
     & $N_{explor}$ & 100 & Population size (exploration) \\ \bottomrule
    \end{tabular}
\end{table}

%---------------------- Performance of HYGO -------------------
\subsection{Performance Evaluation of \HYGO as parametric optimiser} \label{ss:LGA_analytical_kfold}
Evaluating the performance of a stochastic optimisation method necessitates rigorous $k$-fold statistical analysis. A set of 15 analytical benchmark functions featuring diverse and challenging topologies was used to assess \HYGO’s behaviour. To probe scalability, some functions were optimised in low- and high-dimensional variants, resulting in 20 distinct scenarios. The study compares classical GA, \HYGO, DE, PSO, L-SHADE, and CMA-ES (the last four only for high-dimensional cases), with a limit of 5000 function evaluations per run. Each scenario was repeated for 50 independent runs with randomised initialisations, ensuring robust statistical coverage and mitigating initialisation bias. Summary results are presented in \autoref{tab:lga_analytical_results}, with function definitions and search domains listed in \autoref{tab:analytical_functions}.
\begin{table*}
    \centering
    \caption{Analytical Benchmark Functions: Mathematical Definitions and Parametric Search Domains}
    \label{tab:analytical_functions}
    \resizebox{\textwidth}{!}{%
    \begin{tabular}{|l|l|l|}
    \hline
    \textbf{Function} & \textbf{Mathematical Formulation} & \textbf{Search Space} \\ \hline
    \makecell{\centering Ackley} & $f(\mathbf{x}) \; = \; -20 \exp\left(-0.2 \sqrt{\frac{1}{n} \sum_{i=1}^{n} x_i^2}\right) - \exp\left(\frac{1}{n} \sum_{i=1}^{n} \cos(2\pi x_i)\right) + e + 20$ & $x_i \; \in \; [-5, 5]$ \\ \hline
    \makecell{\centering Beale} & $f(x_1, x_2) \; = \; (1.5 - x_1 + x_1 x_2)^2 + (2.25 - x_1 + x_1 x_2^2)^2 + (2.625 - x_1 + x_1 x_2^3)^2$ & $x_1, x_2 \; \in \; [-4.5, 4.5]$ \\ \hline
    \makecell{\centering Booth} & $f(x_1, x_2) \; = \; (x_1 + 2x_2 - 7)^2 + (2x_1 + x_2 - 5)^2$ & $x_1, x_2 \; \in \; [-10, 10]$ \\ \hline
    \makecell{\centering Bukin N.6} & $f(x_1, x_2) \; = \; 100 \sqrt{|x_2 - 0.01x_1^2|} + 0.01 |x_1 + 10|$ & $x_1 \; \in \; [-15, -5], \; x_2 \; \in \; [-3, 3]$ \\ \hline
    \makecell{\centering Easom} & $f(x_1, x_2) \; = \; -\cos(x_1)\cos(x_2)\exp\left(-((x_1-\pi)^2 + (x_2-\pi)^2)\right)$ & $x_1, x_2 \; \in \; [-100, 100]$ \\ \hline
    \makecell{\centering Eggholder} & $f(x_1, x_2) \; = \; -(x_2 + 47) \sin\left(\sqrt{|x_1/2 + (x_2 + 47)|}\right) - x_1 \sin\left(\sqrt{|x_1 - (x_2 + 47)|}\right)$ & $x_1, x_2 \; \in \; [-512, 512]$ \\ \hline
    \makecell{\centering Goldstein-Price} & $f(x_1, x_2) \; = \; \left[1 + (x_1 + x_2 + 1)^2(19 - 14x_1 + 3x_1^2 - 14x_2 + 6x_1x_2 + 3x_2^2)\right]$ & $x_1, x_2 \; \in \; [-2, 2]$ \\ 
     & $\times \left[30 + (2x_1 - 3x_2)^2(18 - 32x_1 + 12x_1^2 + 48x_2 - 36x_1x_2 + 27x_2^2)\right]$ &  \\ \hline
    \makecell{\centering Himmelblau's} & $f(x_1, x_2) \; = \; (x_1^2 + x_2 - 11)^2 + (x_1 + x_2^2 - 7)^2$ & $x_1, x_2 \; \in \; [-6, 6]$ \\ \hline
    \makecell{\centering Holder Table} & $f(x_1, x_2) \; = \; -|\sin(x_1)\cos(x_2)\exp\left(|1 - \sqrt{x_1^2 + x_2^2}/\pi|\right)|$ & $x_1, x_2 \; \in \; [-10, 10]$ \\ \hline
    \makecell{\centering Levi N.13} & $f(x_1, x_2) \; = \; \sin^2(3\pi x_1) + (x_1-1)^2 \left(1 + \sin^2(3\pi x_2)\right) + (x_2-1)^2 \left(1 + \sin^2(2\pi x_2)\right)$ & $x_1, x_2 \; \in \; [-10, 10]$ \\ \hline
    \makecell{\centering Matyas} & $f(x_1, x_2) \; = \; 0.26(x_1^2 + x_2^2) - 0.48x_1x_2$ & $x_1, x_2 \; \in \; [-10, 10]$ \\ \hline
    \makecell{\centering Sphere} & $f(\mathbf{x}) \; = \; \sum_{i=1}^{n} x_i^2$ & $x_i \; \in \; [0,2]$ \\ \hline
    \makecell{\centering Rastrigin} & $f(\mathbf{x}) \; = \; 10n + \sum_{i=1}^{n} \left[x_i^2 - 10 \cos(2\pi x_i)\right]$ & $x_i \; \in \; [-5.12, 5.12]$ \\ \hline
    \makecell{\centering Rosenbrock} & $f(\mathbf{x}) \; = \; \sum_{i=1}^{n-1} \left[100(x_{i+1} - x_i^2)^2 + (x_i - 1)^2\right]$ & $x_i \; \in \; [-5, 5]$ \\ \hline
    \makecell{\centering Styblinski-Tang} & $f(\mathbf{x}) \; = \; \frac{1}{2} \sum_{i=1}^{n} \left(x_i^4 - 16x_i^2 + 5x_i\right)$ & $x_i \; \in \; [-5, 5]$ \\ \hline
    \end{tabular} }
\end{table*}
\begin{table*}
    \centering
    \caption{Comparative performance of GA and \HYGO on the analytical benchmark functions from \autoref{tab:analytical_functions} (some in both 2D and 25D cases). For each function, the table presents the percentage of successful convergence, average number of generations, evaluations, and best achieved cost over 50 independent runs with randomised initialisation. Bold values indicate the superior algorithm for each metric. The bottom row summarises the frequency each method outperforms the other in each category, providing a global assessment of robustness and efficiency.} \label{tab:lga_analytical_results}
    \begin{tabular}{|c|c|c||c|c||c|c||c|c|}
    \hline
    \multirow{2}{*}{Name} & \multicolumn{2}{c||}{Convergence \%} & \multicolumn{2}{c||}{Generations} & \multicolumn{2}{c||}{Evaluations} & \multicolumn{2}{c|}{Best Cost} \\ \cline{2-9} 
                           & GA           & \HYGO          & GA           & \HYGO          & GA           & \HYGO             & GA           & \HYGO          \\ \hline
    Ackley-25D              & 0            & 0             & 50           & 50            & 4946.1 & 5000           & 7.401        & \textbf{5.259}        \\ \hline
    Ackley-2D               & \textbf{100} & 98            & 14.48        & \textbf{5.72} & 1290.3       & \textbf{548.72}   & \textbf{0.002} & 0.054      \\ \hline
    Beale                  & 2            & \textbf{86}   & 49.22        & \textbf{14.08}& 2689.9       & \textbf{1372.7}   & \textbf{0.030}  & 0.107       \\ \hline
    Booth                  & 6            & \textbf{100}  & 48.58        & \textbf{2.82} & 2542.4       & \textbf{264.8}    & 0.125      & \textbf{2.082e-07}    \\ \hline
    Bukin N.6                & 0            & 0             & 50           & 50            & 3706.8 & 4948.2         & 0.124       & \textbf{0.071}      \\ \hline
    Easom                  & 14           & \textbf{96}   & 45.8         & \textbf{11.12}& 2640.1       & \textbf{1077.9}   & -0.660     & \textbf{-0.960}      \\ \hline
    Eggholder              & 4            & \textbf{18}   & 48.54        & \textbf{41.88}& \textbf{2555.6} & 4096.4         & \textbf{-915.620}  & -896.190       \\ \hline
    Goldstein-Price              & 100          & 100           & 17.8         & \textbf{4.92} & 1555.5       & \textbf{468.78}   & 3.000            & 3.000             \\ \hline
    Himmelblaus            & 54           & \textbf{100}  & 36.28        & \textbf{3.62} & 2117.6       & \textbf{340.2}    & 0.004 & \textbf{7.237e-07}    \\ \hline
    Holder Table            & 54           & \textbf{100}  & 34.02        & \textbf{4.14} & 2092.5       & \textbf{390}      & -19.207      & \textbf{-19.209}       \\ \hline
    Levi N.13                & 12           & \textbf{22}   & 45.76        & \textbf{43.94}& \textbf{2512.7} & 4243.2       & \textbf{0.011} & 0.015      \\ \hline
    Matyas                 & 56           & \textbf{100}  & 37.64        & \textbf{3.42} & 2286.6       & \textbf{321.24}   & 9.267e-4 & \textbf{9.127e-09}    \\ \hline
    Sphere-25D             & 0            & 0             & 50           & 50            & 4950.9       & 4909.6   & 0.133      & \textbf{0.0219}      \\ \hline
    Sphere-2D              & 100          & 100           & 13.62        & \textbf{1.76} & 1250.2       & \textbf{151.02}   & 0.000            & 0.000             \\ \hline
    Rastrigin-25D           & 0            & 0             & 50           & 50            & 4735.5 & 4988.1        & 63.989       & \textbf{31.346}        \\ \hline
    Rastrigin-2D            & \textbf{98}  & 80            & \textbf{16.36} & 21.04        & \textbf{1341.6} & 2029.3       & \textbf{0.025} & 0.264       \\ \hline
    Rosenbrock-25D          & 0            & 0             & 50           & 50            & 4950.8       & 4917.7   & 10182.000        & \textbf{193.290}        \\ \hline
    Rosenbrock-2D           & 2            & \textbf{76}   & 49.52        & \textbf{29.14}& \textbf{2710.7}       & 2863.4   & 0.207      & \textbf{0.129}       \\ \hline
    Styblinski-Tang-25D      & 0            & 0             & 50           & 50            & 4950.7      & 5000             & -932.050      & \textbf{-936.580}       \\ \hline
    Styblinski-Tang-2D       & 72           & \textbf{100}  & 30.86        & \textbf{3.4}  & 1884.8       & \textbf{317.94}   & \textbf{-78.331} & -78.332       \\ \hline\hline
    \textbf{Total Wins}    & 2            & \textbf{10}   & 1            & \textbf{13}   & 4           & \textbf{10}       & 6           & \textbf{12}   \\ \hline
    \end{tabular}
\end{table*}

\HYGO consistently outperforms the classical GA across measured metrics in \autoref{tab:lga_analytical_results}. The convergence criterion is strict, demanding identification of the exact global minimum parameter set. \HYGO achieves superior convergence on half of the benchmark functions, often by a substantial margin in the best objective costs. Equivalent convergence is observed on eight functions, although four are high-dimensional cases where the probability of exact convergence is extremely low due to the vast discretised search space ($2^{12}$ points per variable in 25 dimensions). In these instances, mean best cost offers a more realistic metric of performance, and \HYGO delivers notable improvements.

Exceptions include the Ackley, Rastrigin, and Sphere functions in two dimensions, where GA achieves higher convergence rates and better mean best cost; the Sphere function is trivial for both algorithms, given its geometric simplicity. Notably, in functions where convergence is achieved, \HYGO generally requires fewer evaluations and generations than GA, highlighting the efficiency gained by DSM exploitation. Special consideration is required for functions with zero convergence rates, such as Bukin N.6, notable for its complex valley topology densely populated by similar local minima. Caution is advised when interpreting evaluation count metrics in zero-convergence scenarios due to potential re-sampling limits, i.e. repeated individuals that could not be re-generated within the maximum number of attempts. 
\begin{figure*}
    \centerline{\includegraphics[width=0.9\linewidth]{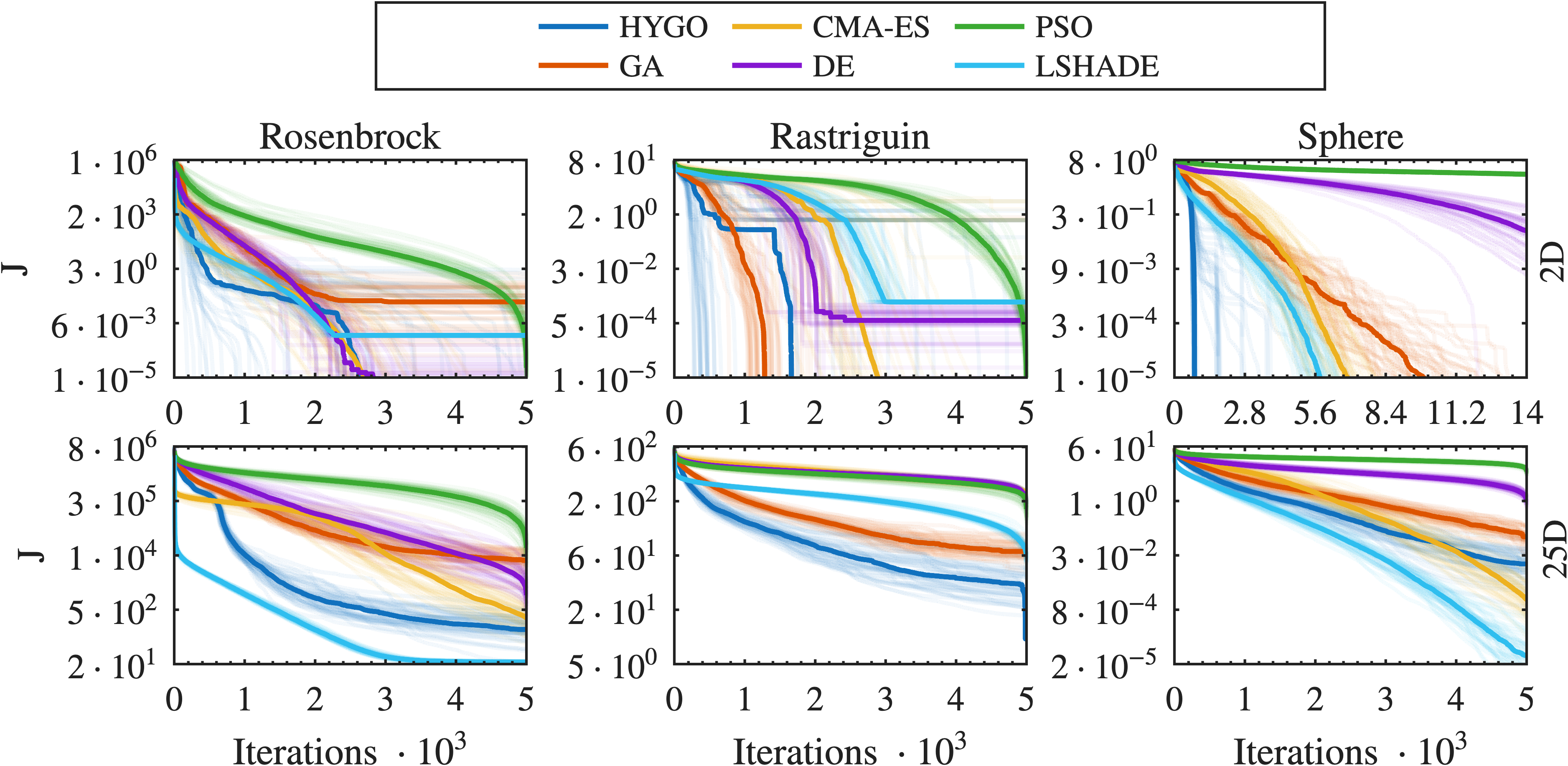}}
    \caption{Convergence comparison of \HYGO, GA, and CMA-ES algorithms on the Rosenbrock, Rastrigin, and Sphere benchmark functions in 2D (top row) and 25D (bottom row). Each panel shows the evolution of the cost function $J$ as a function of the number of iterations across 50 independent runs with randomised initial conditions. The lighter trajectories display individual runs while the darker lines denote the median of all runs. This visualisation highlights both the average performance and variability of each algorithm under different dimensionalities, illustrating convergence rates and robustness in complex landscapes.
    }\label{fig:AnalyticalResults2D25D}
\end{figure*}

Overall, the inclusion of DSM in the optimisation process substantially improves both solution quality and computational efficiency across a broad suite of analytical benchmarks. In terms of cost, \HYGO delivers a superior result in 12 scenarios compared to 6 for GA, while also requiring fewer evaluations (10 versus 4), demonstrating the robustness and advantage of the hybrid strategy. These averages, derived from 50 independent runs per case, further underline the statistical advantage imparted by hybridisation, despite occasional outlier effects from especially favourable or unfavourable initialisations.

%-------- High-dimensional
\subsection{High-dimensionality performance}\label{ss:LGA_analytic_high_d}
The Rosenbrock, Rastrigin, and Sphere functions were selected to interrogate the influence of dimensionality on algorithmic performance. These functions, by virtue of their topological differences, provide archetypes for gradients (Sphere), multi-modality requiring broad exploration (Rastrigin), and landscapes mixing exploration and exploitation demands (Rosenbrock). Notably, GA outperforms or matches \HYGO in certain low-dimensional scenarios, highlighting the importance of both landscape and dimensional scaling.
As visualised in \autoref{fig:AnalyticalResults2D25D} (and further quantitatively analysed in \autoref{anex:Comparison}), each panel presents the evolution of the objective value $J$ as a function of iteration for 50 independent runs per algorithm. The upper row shows 2D cases; the lower row, 25D. Solid lines correspond to and median performances, with lighter lines for individual trajectories, providing insight into both average trends and robustness.

For the 2D Rastrigin function (\autoref{fig:AnalyticalResults2D25D}, middle, top), the classic GA's strong exploratory focus enables it to find the global minimum effectively, outperforming \HYGO, CMA-ES, DE, L-SHADE, and PSO in median performance. DE, L-SHADE and PSO, while competitive initially, stall after approximately $2 \times 10^3$ iterations at suboptimal solutions. In contrast, the classic GA is least effective on the gradient-dominated Sphere function (\autoref{fig:AnalyticalResults2D25D}, top right), in which exploitation is critical. Here, \HYGO converges fastest and achieves the lowest median cost, significantly outperforming all other algorithms.

On the Rosenbrock function (\autoref{fig:AnalyticalResults2D25D}, top left), which demands a combination of global search and exploitation, CMA-ES ultimately achieves the lowest median cost by leveraging path exploitation, enabling it to catch up to \HYGO, which rapidly optimises the function in the early stages. A similar behaviour is shown by DE, whose initial optimisation is slower, but due to its adaptive population capabilities matches CMA-ES and \HYGO in final performance. Furthermore, \HYGO remains highly competitive and demonstrates strong robustness across runs, outperforming PSO in final median quality and comparable results with GA and CMA-ES. L-SHADE and PSO show rapid initial improvement, but their median convergence stalls well above the CMA-ES and \HYGO final costs.

The high-dimensional (25D) results, shown in the lower row, exhibit dampened impact from initial sampling: all algorithms initialise from poorly informed, sparsely distributed populations. Here, in the high-dimensional Sphere, \HYGO and CMA-ES are initially similar, being \HYGO faster at the start (strengthening the claim that geometric-based gradient approximation is more efficient than CMA-ES) and reaching almost at the same time a cost value of $J\approx 10^{-2}$, which is already very low. However, CMA-ES benefits from memory in later iterations, slightly outperforming \HYGO. Both DE and PSO show diminished effectiveness in this high-dimensional setting, trailing \HYGO and CMA-ES due to their inability of exploiting local gradient information and, in the case of PSO, the lack of particle velocity effectiveness in such a high dimensional scenario. The GA lags substantially due to its inherent lack of exploitative capabilities and the significant increase in problem dimensionality, which can no longer be tackled through exploration alone. Lastly, the best-performing algorithm is L-SHADE. The linear reduction in population size and history-based parameter adaptation (similar to the path memory of CMA-ES) enables L-SHADE to maintain a continuous improvement leveraging the evident gradient of the function.

For the Rastrigin benchmark, both GA and CMA-ES struggle with exploration in the vast search space, and only \HYGO can maintain an effective balance, lowering the objective more rapidly and to a greater degree through geometric exploitation once promising regions are found. The large number of minima slows down the evolution of CMA-ES. Moreover, due to the large number of parameters, the different minima are more challenging to exploit because finding the gradient direction is more complex and finding the basin of the global minimum is no longer enough for convergence. Thus, the ability to utilise the gradient information once the basin of the global minimum is found gains importance, leading to \HYGO excelling in the 25-dimensional Rastrigin function. In this highly non-convex, multimodal landscape, L-SHADE experiences significant performance degradation due to the dense distribution of local optima. The algorithm is not able to effectively leverage its history-based parameter adaptation and linear population size reduction. Consequently, L-SHADE lacks the exploration mechanisms to find new regions of the search space in the later stages of optimisation, leading to premature convergence. Similarly, both PSO and DE struggle under these conditions since the particle movement in PSO and the population-level adaptation in DE are unable to properly navigate the function's topology.

In high-dimensional Rosenbrock, \HYGO and CMA-ES again show strong late-stage convergence, with GA falling behind for most of the optimisation horizon. The median performance of DE and PSO is competitive with GA but significantly worse than the best median achieved by \HYGO or CMA-ES due to their less effective exploration mechanisms compared to the mutation operation. The challenge of approximating a high-dimensional gradient is also appreciated in the 25D Rosenbrock function, where the CMA-ES algorithm performs worse than GA in intermediate evaluation numbers. However, path exploitation enables the CMA-ES to achieve performance metrics similar to \HYGO's in later stages, where the geometry is very shallow. Similarly, the more simple history-based adaptation of L-SHADE enables it to avoid the vanishing gradient phenomenon that likely slows \HYGO in the latter stages, outperforming all algorithms in this function.

Overall, these results demonstrate that high-dimensional settings increase the relative value of effective local exploitation once regions of interest are identified, explaining the superior or comparable performance of \HYGO against established methods, including DE, CMA-ES, and PSO algorithms. Notably, \HYGO consistently delivers a robust trade-off between exploration and exploitation, yielding competitive or superior convergence profiles as dimensionality increases. The primary strength of \HYGO lies in its topological robustness, providing consistent and reliable performance across a wide array of objective function landscapes and complex search spaces. While L-SHADE is recognised as a sophisticated, advanced metaheuristic capable of high performance, it demonstrates notable sensitivity in high-dimensional, non-convex scenarios, as for the Rastrigin benchmark.

%------------------------------------------------------
%--------------- The Damped Landau -------------------
%------------------------------------------------------
\section{Functional learning benchmark: The Damped Landau Oscillator}\label{s:GP_validation}
\begin{figure}
    \centerline{\includegraphics[width=0.45\linewidth]{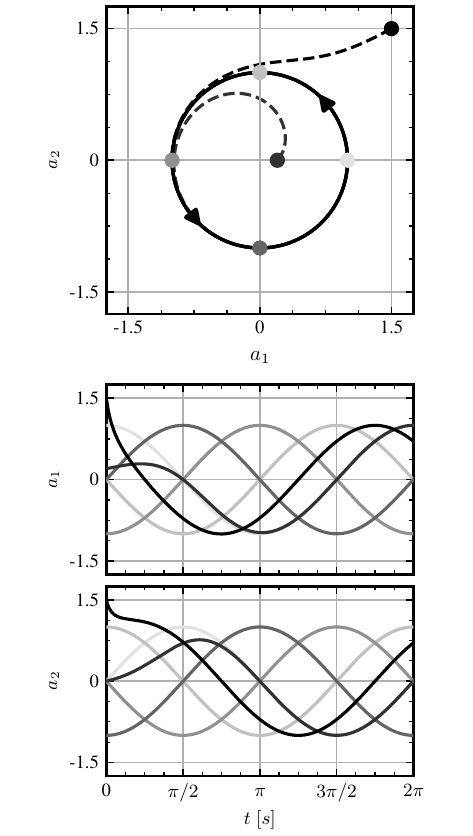}}
    \caption{Phase space and time evolution of the undamped Landau oscillator for six representative initial conditions $(a_1,a_2)_{t=0} = \left[\left(1,0\right),\left(-1,0\right),\left(0.2,0\right),\left(0,1\right),\left(0,-1\right),\left(1.5,1.5\right)\right]$. Left: Orbits in the $(a_1,a_2)$ phase plane illustrating stable periodic motion on the unit circle. Right: Time histories of $a_1$ (top) $a_2$ (bottom) for each initial state, showing sustained oscillatory dynamics. Each curve corresponds to one initial condition, highlighting solution periodicity and dependence on initialisation.}\label{fig:Landau_no_actuation}
\end{figure}

The enhanced LGP optimiser is benchmarked on the Landau oscillator problem. It is a two-dimensional system displaying exponential growth and nonlinear saturation that has already been employed in \citet{CornejoMaceda2022book}. The relative simplicity of the system makes it an ideal benchmark for evaluating the ability of LGP to evolve effective control strategies and assessing the impact of Subplex local search enrichment on optimisation performance. Moreover, the Landau oscillator captures the oscillatory dynamics of von Kármán vortex shedding behind a cylinder \citep{Luchtenburg2009Landau}, making it a relevant proxy for fluid flow control scenarios. The governing equations of the Landau oscillator are:
\begin{equation}
    \begin{cases}
        \dot{a}_1 = \left(1 - a_1^2 - a_2^2\right) a_1 - a_2, \\
        \dot{a}_2 = \left(1 - a_1^2 - a_2^2\right) a_2 + a_1 + b(a_1, a_2), \\
    \end{cases}
\end{equation}
where $b(a_1,a_2)$ represents the control input. Without control ($b=0$), the system exhibits a stable periodic orbit on the unit circle, with solutions rotating counterclockwise indefinitely. \autoref{fig:Landau_no_actuation} illustrates six representative initial conditions rapidly entering the stable limit cycle. 

The optimisation goal is to stabilise the oscillator at the origin $(a_1,a_2)_{t=t_f} = (0,0)$ as fast as possible with minimal control effort. The cost function $J$ combines a stabilisation term $J_a$ (the mean squared distance to the origin) and an actuation penalty $J_b$ (the mean control energy), as defined in \autoref{eq:landau_cost_function}. Formally, the proposed optimisation problem is multi-objective. Since \HYGO is a single-objective algorithm, scalarisation is achieved by selecting an appropriate weight $\gamma$ based on the following single-objective definition of the cost function:
\begin{equation}\label{eq:landau_cost_function}
\begin{split}
    J = &J_a + \gamma J_b = \\=&\left(\frac{1}{t_{end}} \int_0^{t_{end}} a_1^2 + a_2^2\; dt\right) +\\&+ \gamma \left(\frac{1}{t_{end}} \int_0^{t_{end}} b(a_1, a_2)^2\; dt\right)
\end{split}
\end{equation}

\begin{figure*}[t]
    \centering
    \includegraphics[width=0.9\linewidth]{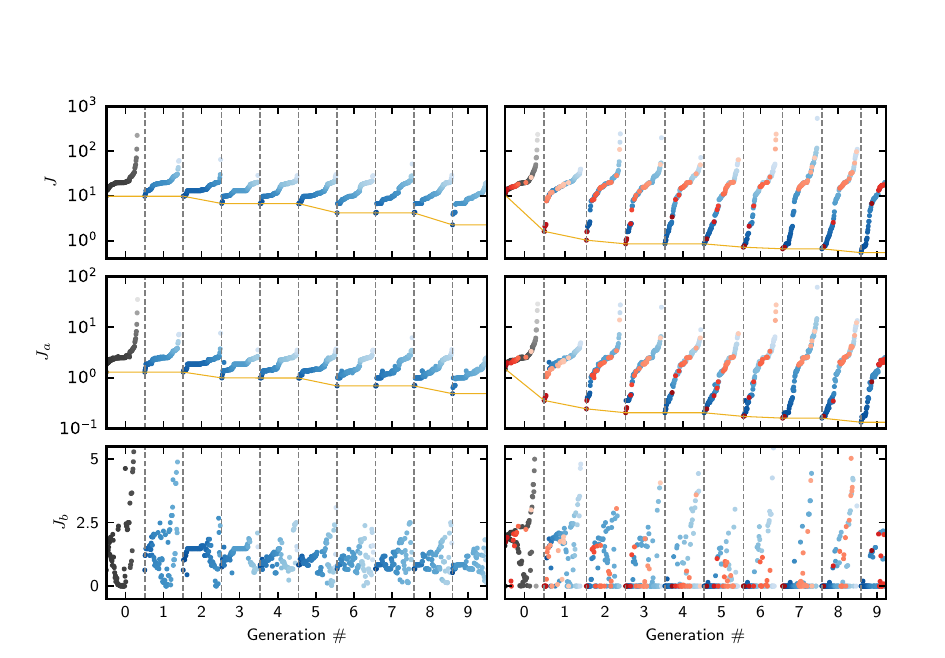}
        \caption{Evolution of the total cost $J$ (top row), stabilisation cost $J_a$ (middle), and the penalisation $J_b$ (bottom) through the generations for a GP optimisation of the damped Landau oscillator, comparing cases without (left) and with (right) Subplex local search enrichment. Each dot represents an individual, coloured by origin: black for random initialisation, blue for genetic operations, and red for Subplex-generated individuals; darker dots indicate lower cost. The yellow line traces the best cost found per generation. }\label{fig:Landau_J}
\end{figure*}
with $t_{end}=20T$, and $T=2\pi$ corresponding to the oscillator's fundamental period given that the angular frequency of the uncontrolled oscillator is 1.

In this setup, the control input influences only the second equation, simplifying the optimisation to a single control law. Both state variables $(a_1,a_2)$ serve as sensor inputs to the control law, ensuring adaptability to instantaneous system states. While more complex problems may require historical sensor values (e.g., delayed coordinates $a_i(\tau-kT/4)$ for $k=0,\dots,4)$, as in \citet{castellanos2022GP}), the Landau oscillator’s simplicity allows effective optimisation without this complexity, avoiding unnecessary expansion of the solution space which can hamper convergence.

\begin{figure*}[t]
    \centering
    \includegraphics[width=0.93\linewidth]{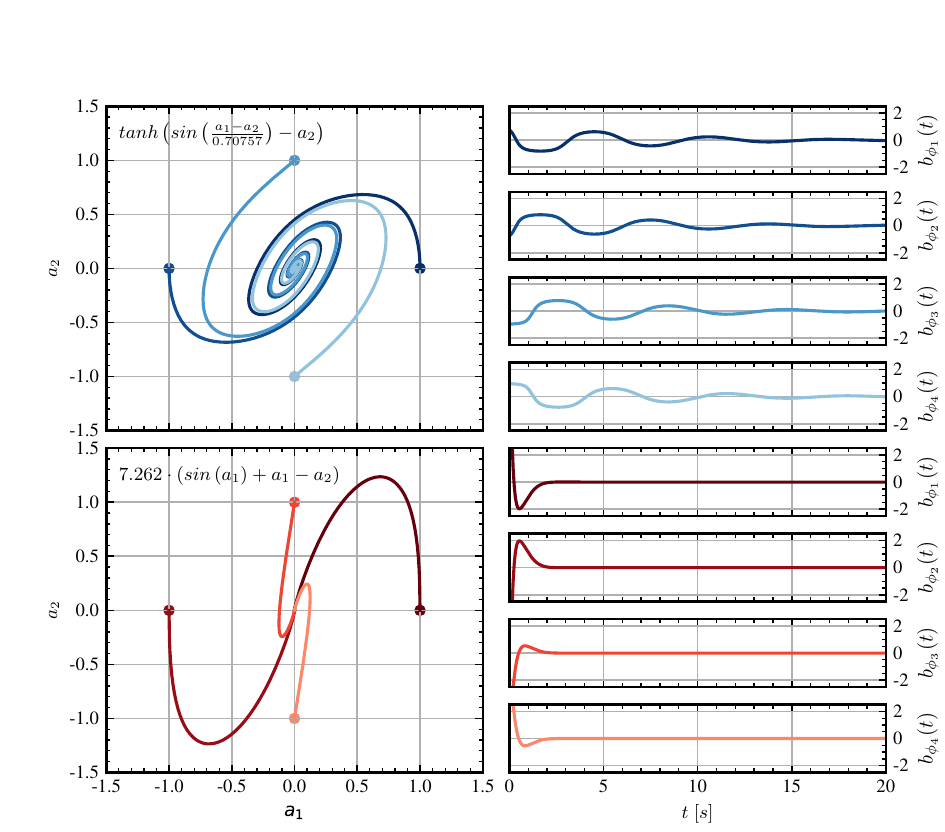}
        \caption{Comparison of optimal control laws discovered for the damped Landau oscillator via genetic programming, without (top) and with (bottom) Subplex enrichment. Left: Phase space trajectories show the system state $(a_1,a_2)$ evolving from multiple initial conditions under the learned feedback. The control law expression for the best individual is depicted in each panel. Right: Corresponding control actuation $b_{\phi_i}(t)$ applied over time for each initial state, $\phi_i$. Subplex-enriched optimisation yields faster and more direct stabilisation to the origin, demonstrated by tighter and less oscillatory trajectories and sparser control input.}\label{fig:Landau_optima}
\end{figure*}

To assess the impact of Subplex local search enrichment, two optimisations were performed, with and without Subplex exploitation, using the average cost across four initial conditions $(a_1,a_2)_{t=0} = \left\{\left(1,0\right),\left(-1,0\right),\left(0,1\right),\left(0,-1\right)\right\}$ for generalisation. The evolution of the optimisation process is illustrated in \autoref{fig:Landau_J}. Each optimisation spanned 10 generations of 100 individuals, with the hybrid approach allocating 80 individuals to the explorative phase and 20 to Subplex-based exploitation. Initialisation consisted of randomly generated individuals with several instructions ranging between 5 and 35 after intron elimination. Thereafter, the minimum number of instructions was 2. An increased number of minimum instructions enabled initialising the individuals with complex control laws, allowing the algorithm to simplify them. After some analysis, this proved the best approach since a low number of instructions leads to constant control laws. However, initialising the optimisation with fewer instructions forces complex operations (including trigonometric functions, sensors\dots). Registers included a zero-initialised output register to avoid bias, two sensors for the current state, a randomly initialised memory register, and two constant registers holding random values in the range $[0,1]$. Regarding the subplex, no definitive recommendation exists for Subplex size due to the problem’s effective infinite dimensionality. Empirical testing indicated that selecting the top 10 individuals (10\% of the population) balances fittest retention with control law variability. Larger Subplex subpopulations tend to increase output law complexity, adversely affecting interpolation efficiency. Nevertheless, this parameter highly depends on user experience and the targeted optimisation problem.

\autoref{fig:Landau_J} depicts the evolution of total cost $J$, stabilisation term $J_a$, and control penalisation $J_b$. The optimisation is controlled by the stabilisation term $J_a$ (due to the low value set to the weighting parameter $\gamma$), leading to a very similar progress of both $J$ and $J_a$. Both algorithms demonstrate an ability to reduce $J_a$ and $J_b$ over successive generations, adapting control laws to trade off effectiveness (stabilisation) and efficiency (control sparseness). However, the trajectory of cost evolution is markedly different between approaches. Notably, \HYGO’s Subplex enrichment results in rapid, early-stage reductions in both cost terms, evident as a step-change improvement early in the hybrid run, whereas pure LGP produces steady but incremental gains.

A critical qualitative distinction emerges in how the algorithms target the conflicting goals embedded in $J_a$ and $J_b$. While both approaches reduce $J_a$ by driving the system rapidly towards the origin, \HYGO more effectively minimises $J_b$, leveraging Subplex-generated individuals that achieve stabilisation using sharp, short-duration bursts of control. This results in actuation that is highly efficient in time, since the control input is intense but localised, minimising its integral, and thus achieves lower overall cost per the objective definition.

\begin{figure*}[t]
    \centering
    \includegraphics[width=0.93\linewidth]{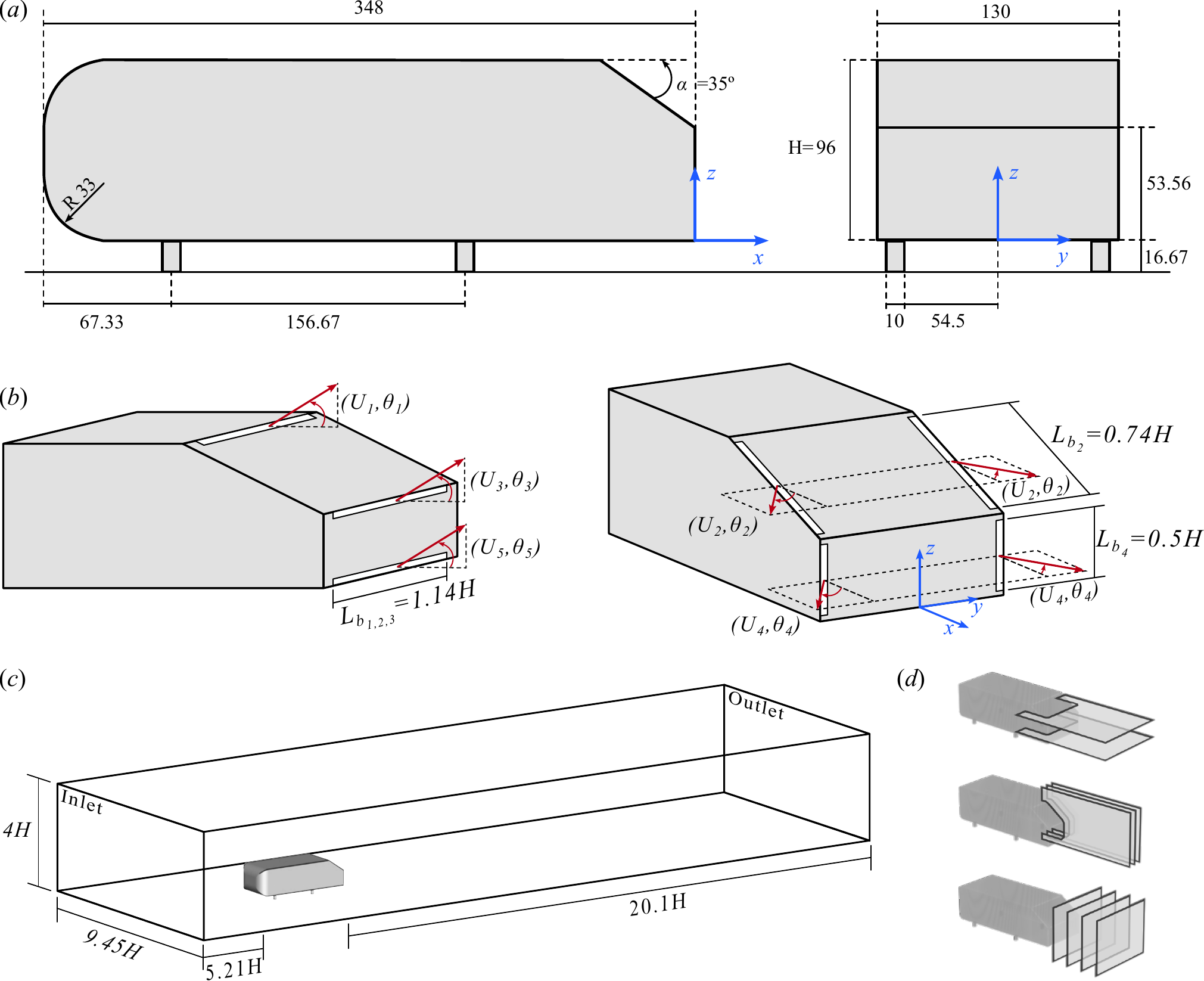}
    \caption{Ahmed body geometry, actuator layout, and computational domain for drag-reduction flow control studies. (a) Side and rear views of the Ahmed body model ($H = 96$), showing all relevant dimensions and axes. (b) Deployment and blowing direction (arrows, defined by actuation speed $U_i$ and angle $\theta_i$) of the five slot actuator groups on the rear window and base. Angles $\theta_i$ are positive when directed outward (right) or upward (left); actuator spanwise lengths $L_{b_i}$  are defined relative to $H$. (c) Computational domain for CFD simulation, locating the Ahmed body within its wind tunnel environment. (d) Schematics of visualisation planes and field of view for visualisation purposes in \autoref{fig:Plano_xy}, \ref{fig:Plano_xz}, and \ref{fig:Plano_yz}. Panels (a,b,c) are adapted with permission from \citet{Li2022JFM_EGM}}.  
    \label{fig:Ahmed_control}
\end{figure*}

Inspection of the best control laws in \autoref{fig:Landau_optima} further underscores these strategic differences. LGP-only optimisation tends to favour solutions built from complex, oscillatory trigonometric expressions; these drive the system towards the origin but often at the price of extended actuation and oscillation before settling, as seen in the phase space and time histories. In contrast, Subplex-enriched \HYGO rapidly identifies and refines control architectures characterised by larger, direct scaling factors, yielding nearly monotonic, efficient convergence to the desired state with sparser and more purposeful actuation. Analysis reveals that the improvement caused by this Suplex-created group stems from emergent large-scale multipliers generated via allowed arithmetic operations, enabling strong initial control inputs that rapidly reduce oscillations. Although the value variable and constant initial registers were limited between $[0,1]$, summations and divisions allow combining scalar numbers beyond 1.

Eventually, both optimisation strategies are capable of evolving control policies that balance stabilisation speed and energy efficiency, yet \HYGO consistently finds superior solutions. Hybridisation via Subplex local search allows \HYGO not only to escape the incrementalism characteristic of mutation/crossover searches but to discover control laws that target the time-localised intervention ideal for the Landau benchmark. These findings reinforce the value of integrating geometric, gradient-free exploitation into evolutionary control synthesis, especially when sparse, interpretable, and efficient functional solutions are sought.

%----------------------------------------------
%------------- Ahmed ---------------------------
%----------------------------------------------
\section{Control of shedding flows: Drag reduction on Ahmed body}\label{s:furgo}
The previous sections established the efficacy of the proposed hybrid optimisation methodology on controlled analytic benchmarks, enabling a detailed examination of \HYGO's exploration and exploitation capabilities. However, while such benchmarks are valuable for method validation, they do not capture the complexity typical of real-world engineering applications, where the parametric landscape is highly nonlinear, multidimensional, and physically constrained. To demonstrate the practical utility of \HYGO, we now address a flow control benchmark: the minimisation of aerodynamic drag on a canonical Ahmed body. This flow control optimisation leverages the simulation setup of \citet{Li2022JFM_EGM} for direct comparison with the state-of-the-art Explorative Gradient Method (EGM), another leading gradient-free hybrid approach.
\begin{table}
    \caption{Parameters for GA and \HYGO}\label{tab:optimization_params_furgo}
    \centering
    \begin{tabular}{lccc}
    \hline
    \textbf{Hyper-Parameter} & \textbf{GA} & \textbf{\HYGO} & \textbf{\HYGO-Stepped} \\ \hline
    Population Size          & 100   & 11    & 50   \\ 
    Max. Generations         & 10    & 100   & 20   \\ 
    Number of bits           & 8     & 8     & 8    \\ 
    Tournament Size          & 2     & 2     & 2    \\ 
    Tournament Probability   & 100\% & 100\% & 100\%\\ 
    Elitism Individuals      & 1     & 1     & 1    \\ 
    Crossover Points         & 1     & 1     & 1    \\ 
    Crossover Mix            & True  & True  & True \\ 
    Mutation Rate            & 0.2   & 0.075 & 0.075\\ 
    Replication Probability  & 0     & 0     & 0    \\ 
    Crossover Probability    & 80\%  & 55\%  & 55\% \\ 
    Mutation Probability     & 20\%  & 45\%  & 45\% \\ 
    Simplex Offspring        & N/A   & 11    & 11   \\ \hline
    \end{tabular}
\end{table}
\begin{table}
    \caption{Simplex initialisation for the second step of \HYGO-Stepped.}\label{tab:simplex_init}
    \centering
    	\begin{tabular}{ccccccccccc}
    	\hline
        Index & $U_1$ & $U_2$ & $U_3$ & $U_4$ & $U_5$ & $\theta_1$ & $\theta_2$ & $\theta_3$ & $\theta_4$ & $\theta_5$ \\\hline
        1 & 10.25 & 6.0 & 1.25 & 8.25 & 7.5 & 27.5 & 0.0 & 0.0 & 0.0 & 0.0\\
        2 & 10.25 & 0.0 & 0.0 & 0.0 & 0.0 & 58.75 & 0.0 & 0.0 & 0.0 & 0.0\\
        3 & 0.0 & 6.0 & 0.0 & 0.0 & 0.0 & 27.5 & 45.0 & 0.0 & 0.0 & 0.0\\
        4 & 0.0 & 0.0 & 1.25 & 0.0 & 0.0 & 27.5 & 0.0 & 45.0 & 0.0 & 0.0\\
        5 & 0.0 & 0.0 & 0.0 & 8.25 & 0.0 & 27.5 & 0.0 & 0.0 & 45.0 & 0.0\\
        6 & 0.0 & 0.0 & 0.0 & 0.0 & 7.5 & 27.5 & 0.0 & 0.0 & 0.0 & 45.0\\
        7 & 10.25 & 0.0 & 0.0 & 0.0 & 0.0 & -3.75 & 0.0 & 0.0 & 0.0 & 0.0\\
        8 & 0.0 & 6.0 & 0.0 & 0.0 & 0.0 & 27.5 & -45.0 & 0.0 & 0.0 & 0.0\\
        9 & 0.0 & 0.0 & 1.25 & 0.0 & 0.0 & 27.5 & 0.0 & -45.0 & 0.0 & 0.0\\
        10 & 0.0 & 0.0 & 0.0 & 8.25 & 0.0 & 27.5 & 0.0 & 0.0 & -45.0 & 0.0\\
        11 & 0.0 & 0.0 & 0.0 & 0.0 & 7.5 & 27.5 & 0.0 & 0.0 & 0.0 & -45.0\\
    	\hline
    	\end{tabular}
\end{table}

\begin{table*}[t]
\caption{Optimal control parameters of the different optimisation strategies, including the cost value and drag reduction.}
\centering
	\begin{tabular}{r@{\quad\quad}l@{\quad\quad}ccccc}
	\hline
	Case & $C_D$   & \multicolumn{5}{c}{Actuation parameters}  \\
	\cline{3-7}
		~ &  (reduction) & Top & Upper & Middle & Lower & Bottom % & Actuation energy
	\\ \hline %[-4pt] \hline \\begin{equation}-18pt]
	Unforced & \hbox to 3em{\rule{0pt}{0pt}\hfill $0.313$ ($0\%$)} 
	&  --- & --- & --- & --- & --- %& ---
	\\
	GA & \hbox to 5em{\rule{0pt}{0pt}\hfill 0.279 ($10.88\%$)}
	& $31.5$m/s
	& $28$m/s
	& $19$m/s
	& $49$m/s
	& $16$m/s
	%   & $2.47\%$
	\\ & 
	& $-28.8^\circ$ 
	& $-50.4^\circ$
	& $-3.6^\circ$ 
	& $-48.6^\circ$
	& $57.6^\circ$
	\\
 	\HYGO & \hbox to 5em{\rule{0pt}{0pt}\hfill 0.265 ($15.6\%$)}
	& $14.5$m/s
	& $17$m/s
	& $4$m/s
	& $19.5$m/s
	& $4.5$m/s
	%   & $2.47\%$
	\\ & 
	& $-27.5^\circ$ 
	& $-43.2^\circ$
	& $18^\circ$ 
	& $-39.6^\circ$
	& $25.2^\circ$
	\\
	\citet{Li2022JFM_EGM} & \hbox to 5em{\rule{0pt}{0pt}\hfill 0.258 ($17.49\%$)}
	& $21.5$m/s
	& $23.81$m/s
	& $1.8$m/s
	& $25.2$m/s
	& $22.5$m/s
	%   & $2.47\%$
	\\ & 
	& $-27^\circ$ 
	& $-42^\circ$
	& $67^\circ$ 
	& $-44^\circ$
	& $22^\circ$
	\\
        \HYGO (stepped) & \hbox to 5em{\rule{0pt}{0pt}\hfill \textbf{0.249} (\textbf{20.56\%})}
        & $34.5$m/s
    	& $36$m/s
    	& $8.5$m/s
    	& $32.5$m/s
    	& $21$m/s
        \\ &
        &$-30^\circ$
        &$-37.8^\circ$
        &$-72^\circ$
        &$-50.4^\circ$
        &$48.6^\circ$
        \\
	\hline
	\end{tabular}\label{tab:optimisation_results}
\end{table*}

\begin{figure*}
    \centering
    \includegraphics[width=0.85\linewidth]{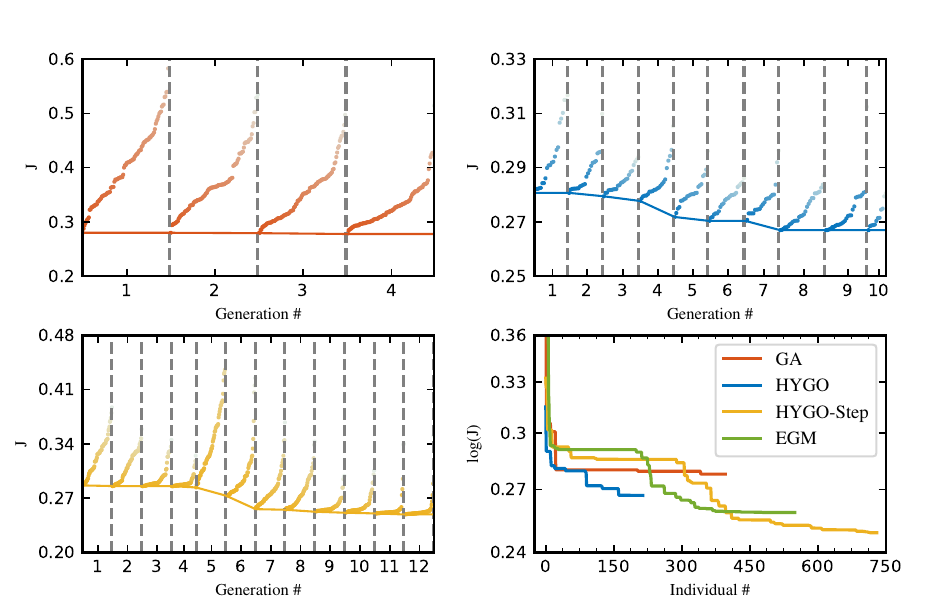}
    \caption{Evolution of the cost function $J$ during the optimisation of drag reduction on the Ahmed body for four strategies. Each panel shows the cost across generations for a different method: GA (top left), \HYGO (top right), and \HYGO-Step (bottom left), while the bottom right plot compares the best cost evolution as a function of individual number for all approaches, including the EGM benchmark (green) from \citet{Li2022JFM_EGM}. Dots correspond to individual evaluations per generation; solid lines trace the minimum cost in each generation. This figure highlights the comparative convergence rates and final performance of the tested algorithms.}
    \label{fig:ahmed_j_evo}
\end{figure*}

\subsection{Ahmed body geometry and Simulation setup}

The reference geometry employed in this study is a 1:3-scale Ahmed body, featuring a slanted rear surface at $\alpha = 35^\circ$. Key model dimensions are $L = 348~\mathrm{mm}$ (length), $W = 130~\mathrm{mm}$ (width), and $H = 96~\mathrm{mm}$ (height). Front edges feature a $0.344 H$ rounding radius, and the body is supported by four cylindrical mounts of $10~\mathrm{mm}$ diameter, and ground clearance of $0.177 H$. A Cartesian coordinate system $(x, y, z)$ with the origin at the midpoint of the rear vertical base's lower edge, aligned with the centreplane (\autoref{fig:Ahmed_control}). Here, $x$, $y$, and $z$ denote streamwise, spanwise, and wall-normal directions, and their velocity components are $u$, $v$, and $w$, respectively. The free-stream velocity is set to $U_\infty = 30~\mathrm{m\,s^{-1}}$.

Flow control is implemented using five groups of steady slot actuators placed along the rear slant and base perimeter, as depicted in \autoref{fig:Ahmed_control}. All slots have a $2~\mathrm{mm}$. The top, middle, and bottom actuators span $109~\mathrm{mm}$; the upper and lower side actuators are $71~\mathrm{mm}$ and $48~\mathrm{mm}$, respectively. The actuation velocities $U_1, \ldots, U_5$ are independent optimisation variables: $U_1$ (upper window edge), $U_3$ (middle), and $U_5$ (bottom of vertical base); $U_2$ and $U_4$ (side actuators). Following \citet{Zhang2018Ahmed} and the same numbering as in velocities, the actuator angles $\theta_i$ are also optimisable, with $\theta_1 \in [-35^\circ, 90^\circ]$ and $\theta_{2,\ldots,5} \in [-90^\circ, 90^\circ]$.

The objective is to minimise the drag coefficient $J = C_D$, that is, to maximise drag reduction, neglecting input power or lift penalties for clarity of comparison. First, a five-dimensional optimisation is conducted, where only the actuation velocities (each bounded not to exceed twice the optimal single-actuator value from \citet{Li2022JFM_EGM}) are variable and angles are fixed at $\theta_i = 0^\circ$ (streamwise blowing). The search space is later expanded to ten parameters by including angular variation as described above.

The numerical and meshing setup matches that in \citet{Li2022JFM_EGM}. The simulation domain, constructed with Ansys ICEM CFD, forms a rectangular wind tunnel (\autoref{fig:Ahmed_control} $(c)$) with dimensions $X_1 \leq x \leq X_2$, $0 \leq z \leq H_T$, and $\vert y \vert \leq W_T/2$, where $X_1 = -5.21 H$, $X_2 = 20.17 H$, $H_T = 4H$, and $W_T = 9.45H$, all exceeding recommended thresholds \citep{Serre2013Ahmed} to minimise boundary effects. The mesh consists of approximately five million hexahedral cells, providing a trade-off between accuracy and computational tractability. Near-wall and actuator slot resolution ($\Delta x^+ = 20$, $\Delta y^+ = 3$, and $\Delta z^+ = 30$) ensures reliable capture of boundary and shear-layer physics.

RANS simulations with the $k$–$\epsilon$ turbulence model are performed in Fluent, employing second-order spatial discretisation and semi-implicit pressure–velocity coupling. For this geometry ($35^\circ$ Ahmed body), RANS has been shown to yield trustworthy estimations \citep{Li2022JFM_EGM} and is thus appropriate for control actuator optimisation.

\begin{figure*}
    \centering
    \includegraphics[width=0.95\linewidth]{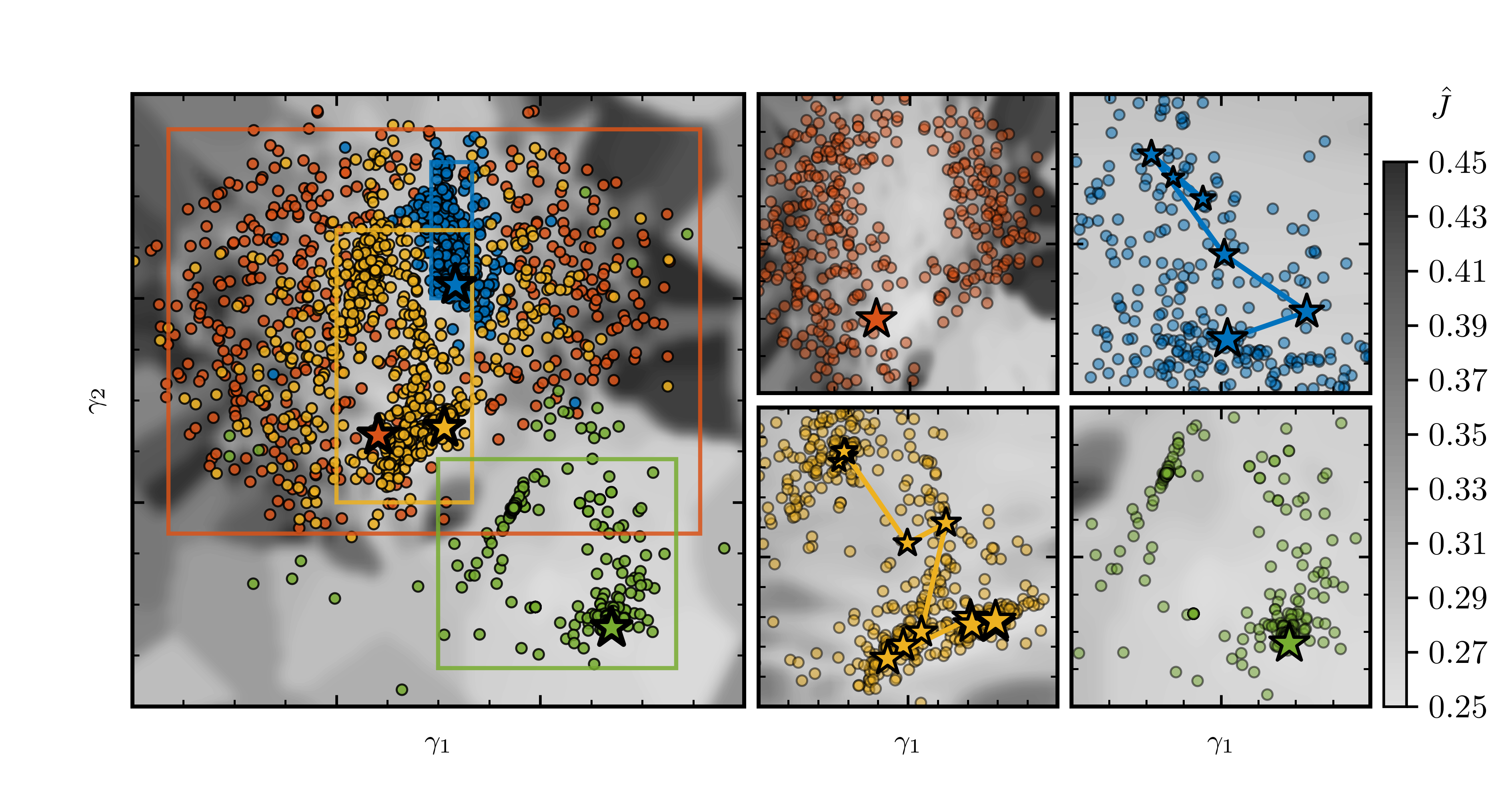}
    \caption[]{Proximity map of the optimisation parameters for all strategies, coloured by cost. The reduced coordinates are normalised.
    \tikz\draw[fill=myorange, draw=black] (0,0) circle (3.5pt); corresponds to GA,
    \tikz\draw[fill=myblue, draw=black] (0,0) circle (3.5pt); to \HYGO,
    \tikz\draw[fill=myyellow, draw=black] (0,0) circle (3.5pt); to \HYGO-stepped, and
    \tikz\draw[fill=mygreen, draw=black] (0,0) circle (3.5pt); to EGM.
    Squares on the general map indicate zoomed-in regions corresponding to each specific optimisation. The zoomed-in subplots illustrate the optimisation path: the best individual from each generation is marked with a star, with the star size increasing over generations (i.e., later generations are represented by larger stars). }
    \label{fig:proximity_ahmed}
\end{figure*}

%---------------- Optimisation  results Ahmed ----------------
\subsection{Optimisation results and data analysis} \label{ss:furgo_results}

The performance of \HYGO is systematically benchmarked against three alternative optimisation frameworks, all evaluated under an identical simulation protocol. The comparison encompasses: (i) the Explorative Gradient Method from \citet{Li2022JFM_EGM}, serving as a reference for local exploitation-driven search; (ii) a conventional Genetic Algorithm with no hybridisation or local search; and (iii) two configurations of DSM-augmented \HYGO: a direct approach optimising all ten control variables simultaneously, and a \emph{stepped} variant in which only the actuation velocities $U_i$ are optimised initially, followed by a second phase where actuation angles $\theta_i$ are introduced. All strategies employ random initial populations to mitigate sampling bias. For the stepped run (\HYGO-stepped), the transition to the full parameter space is seeded by an 11-individual simplex, constructed about the velocity-optimal solution as summarised in \autoref{tab:simplex_init}). 

Key algorithmic and hyper-parameter choices are summarised in \autoref{tab:optimization_params_furgo} for all genetic-based optimisers. A comprehensive summary of the optimisation histories is illustrated in \autoref{fig:ahmed_j_evo}. In all cases, optimisations terminate either upon exhausting the maximum generations (\autoref{tab:optimization_params_furgo}) or earlier if no improvement is observed across four successive generations. This approach provides a balanced assessment of convergence rates and stagnation risks for each algorithm.

The basic GA yielded an initial improvement, attaining a $10.9\%$ reduction in drag compared to the reference value $C_D^0 = 0.3134$, yet failed to realise further gains beyond the best individual in its initial population. The evolutionary process stalled prematurely, evidenced by both early algorithmic termination and broad dispersion of individuals across the parameter space (\autoref{fig:ahmed_j_evo}, orange). This result underscores the limitations of relying solely on stochastic genetic operations for high-dimensional, multimodal fluid-structure optimisation, where complex flow responses may easily trap the process in suboptimal configurations that are unreachable by simple crossover or mutation.

In contrast, \HYGO exhibited a markedly different learning process (\autoref{fig:ahmed_j_evo}, blue). Incorporation of the DSM enabled more consistent improvement across generations, with a final drag reduction of $15.6\%$. The synergistic use of DSM was particularly impactful in early and intermediate stages, with DSM-generated offspring regularly surpassing those from GA-only operations, consistent with literature identifying the benefits of local exploitation for complex design spaces. Nonetheless, as convergence progressed, the algorithm showed signs of diminishing diversity and local entrapment, as reflected in the clustering of individuals within a single region of parameter space in \autoref{fig:proximity_ahmed}, coloured in blue. This concentration is symptomatic of reduced genome variability, which can stymie further exploitation of alternative minima. The hybrid framework’s exploration-exploitation interplay is most clearly highlighted during the first five generations. DSM was capable of rapidly descending to promising local optima, while the broader genetic search provided periodic discoveries of novel attraction basins, events visible as abrupt reductions in the best cost trajectory in \autoref{fig:ahmed_j_evo} blue line. Importantly, once the genetic operations identified a superior region, DSM was able to efficiently refine these new candidates, demonstrating the effectiveness of alternated exploration and exploitation mechanisms.

The stepped \HYGO variant draws direct inspiration from the structured approach of \citet{Li2022JFM_EGM}, introducing an initial five-dimensional subproblem centred only on the actuation speeds. This dimensional reduction is not merely a heuristic for convergence acceleration, but is mathematically motivated: by initially decoupling less influential angular variables, the algorithm reduces the prevalence of spurious local minima linked to variable interactions. The resulting optimisation landscape in this stage has a gentler topology, increasing the probability of successfully locating regions leading to global optimum. Following convergence in this subspace, the full parametric search is reinitiated with an affine simplex sampling (\autoref{tab:simplex_init}), restoring geometric diversity and maximising potential for effective local exploitation by DSM. The stepped strategy proved decisively more effective, attaining a $20.5\%$ drag reduction, outperforming both the best direct ten-dimensional \HYGO and the benchmarked EGM.

Comparative analysis of proximity maps in \autoref{fig:proximity_ahmed} reveals fundamental differences in search behaviour. The GA’s individuals scatter widely (orange), never converging to the most promising region. In contrast, \HYGO’s direct approach (blue) converges rapidly but at the cost of diversity, with limited ability to escape local traps. Conversely, \HYGO-stepped (yellow) achieves a more thorough exploration phase, with early trajectories traversing multiple minima before settling into a lower-cost region. This two-stage protocol leverages both global structure and local refinement, as seen in the sequential contraction of individual spread and ultimate cost function minimisation. Similarly, EGM (green) converges around two distinct local minima, which it exploits efficiently; although its exploration capability is more limited, as indicated by the relatively localised distribution of its individuals in the parameter space.

\begin{figure*}[t]
    \centering
    \includegraphics[width=0.95\linewidth]{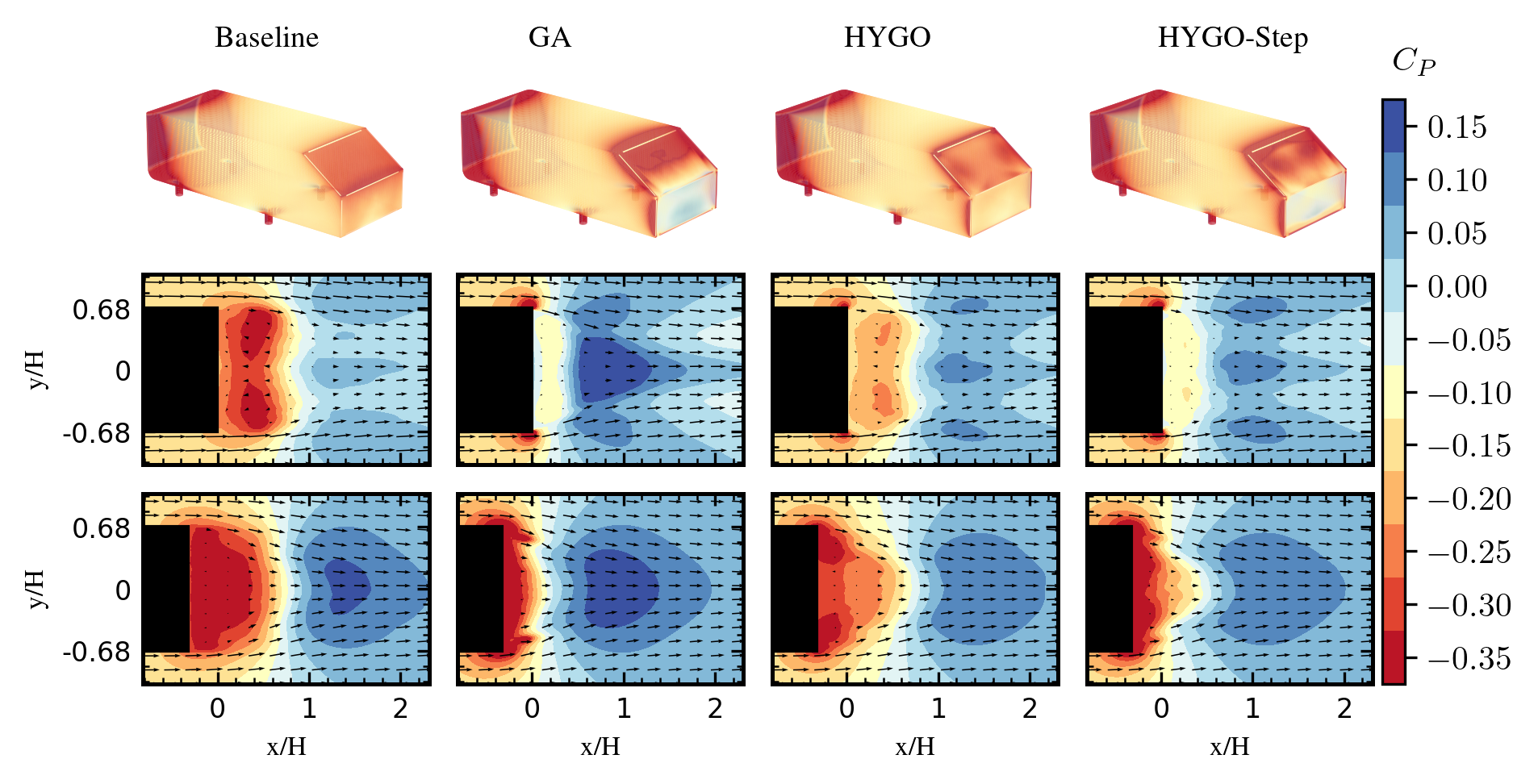}
    \caption{Top view for the surface distribution of the pressure coefficient, $C_p$, on the Ahmed body. Columns, from left to right, represent the reference (baseline), the solution obtained with the genetic algorithm (GA), \HYGO, and \HYGO-stepped methodologies. The first row shows the full‐surface $C_p$ contours, while the second and third rows display $x-y$ pressure fields at the non-dimensional heights $z/H = 0.28$ and $0.785$, respectively. A common colour scale referenced to the free-stream static pressure is applied to every panel to enable quantitative comparison.}
    \label{fig:Plano_xy}
\end{figure*}

Examining the control parameters at the identified optimum listed in \autoref{tab:optimisation_results}, the most successful strategies (\HYGO-stepped and EGM) are found to employ greater actuation authority, in the form of higher velocity magnitudes, while maintaining similar actuation angles required for targeted flow manipulations. Interestingly, the solution found by \HYGO-stepped appears to be a refined version of the one identified by EGM: the actuation angles are similar, but the velocity magnitudes are lower. A key distinction, however, lies in the third actuator. While EGM directs this jet upwards, \HYGO-stepped reverses its direction, potentially offering more direct interfacing with the recirculation bubble. This subtlety may underpin the superior drag reduction achieved and explains the spatial separation between the distinct optima identified by each strategy in the solution proximity map in \autoref{fig:proximity_ahmed}.

In summary, these results elucidate not only the algorithmic advantage conferred by hybrid and staged global–local optimisation, but also provide insight into the physical mechanisms prioritised by algorithmic search: high-authority, tailored actuation strategies emerge as the dominant solution features for robust drag reduction in this geometry.

%---------------- Fluidic comment ------------------
\subsection{Flow-physics interpretation} \label{ss:furgo_results2}
\begin{figure*}[t]
    \centering
    \includegraphics[width=0.95\linewidth]{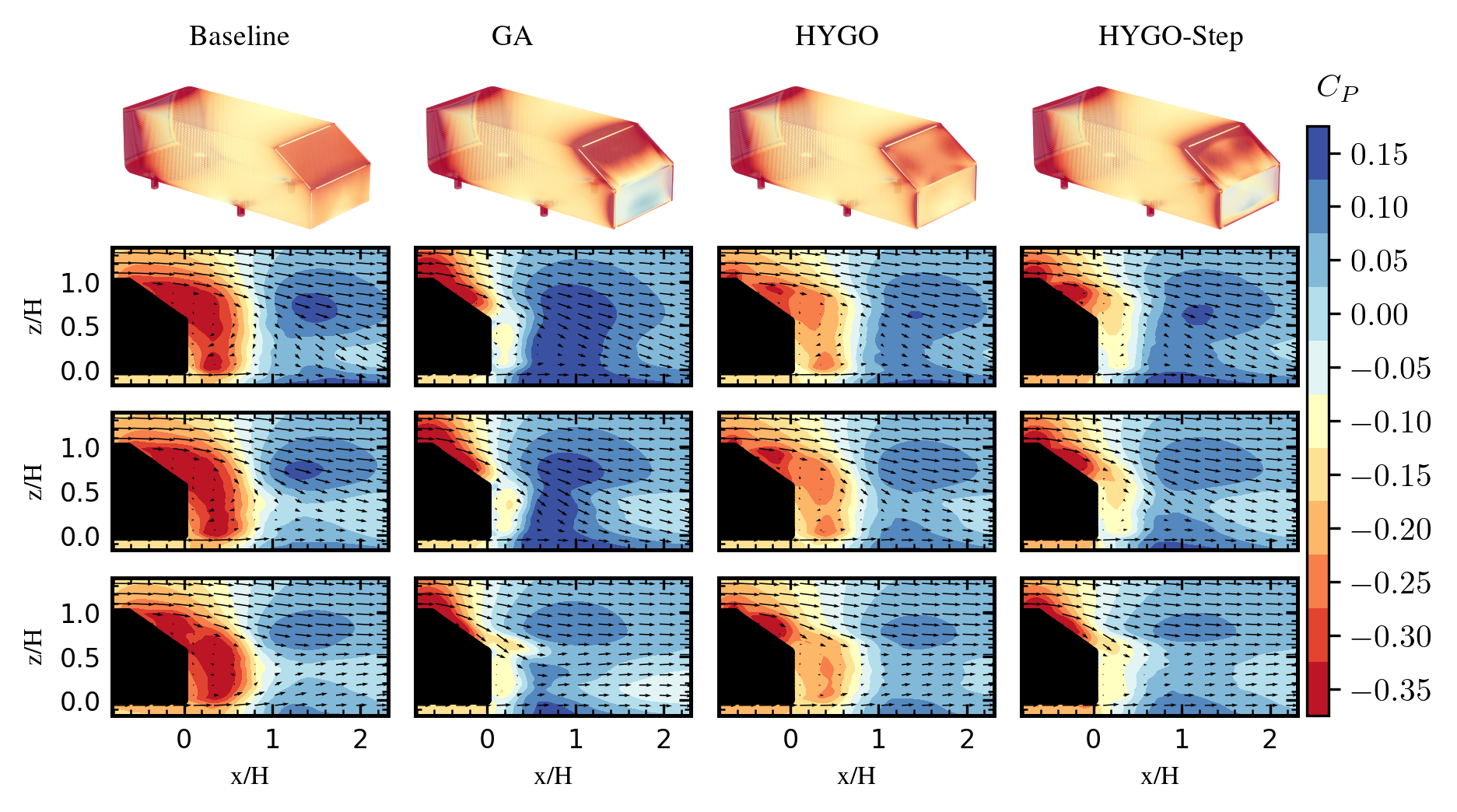}
    \caption{Side view for the surface distribution of the pressure coefficient, $C_p$, on the Ahmed body. Columns, from left to right, represent the reference (baseline), the solution obtained with the genetic algorithm (GA), \HYGO, and \HYGO-stepped methodologies. The first row shows the full‐surface $C_p$ contours, while the second, third, and fourth rows present $x-z$ pressure fields at the spanwise stations $y/H = 0$, $0.225$, and $0.45$, respectively. A common colour scale referenced to the free-stream static pressure is applied to every panel to enable quantitative comparison.}
    \label{fig:Plano_xz}
\end{figure*}

\begin{figure*}[t]
    \centering
    \includegraphics[width=0.9\linewidth]{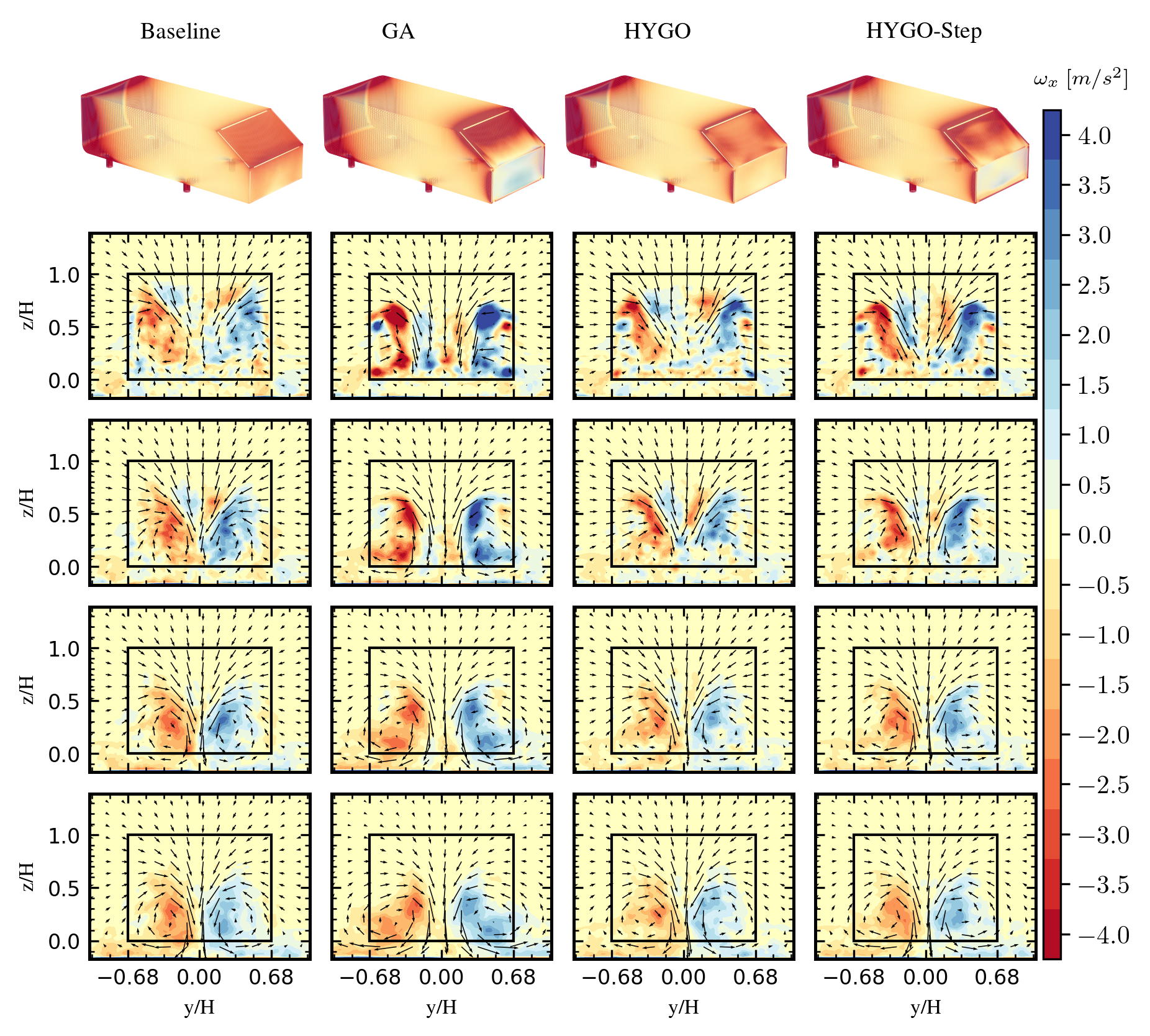}
        \caption{Streamwise vorticity maps for the fittest individuals of each optimisation approach compared to the baseline (no actuation) case at fixed streamwise distances, $x/H$, from the Ahmed body. The highlighted \textbf{black} rectangle represents the Ahmed body's base profile. The first row shows the pressure coefficient distribution on the surface of the Ahmed body in the same scale as \autoref{fig:Plano_xz}. The following rows display the streamwise vorticity maps at $x/H = 0.5$, $1.0$, $1.5$, and $2.0$, respectively. A consistent colour scale is applied across all vorticity maps for clear comparison of the flow structures.}
    \label{fig:Plano_yz}
\end{figure*}

The optimisation exercises not only yield quantitative improvements in drag reduction but also manifest as pronounced alterations to the wake dynamics, as illustrated in \autoref{fig:Plano_xy}–\autoref{fig:Plano_yz}. These figures allow a direct assessment of the flow physics supporting each optimisation outcome, thus assessing the mechanisms via which the algorithms achieve, or fail to achieve, substantive control authority. These figure includes a three-dimensional representation of the pressure distribution (pressure coefficient $C_P$) on the body’s surface, and in the planes highlighted in \autoref{fig:Ahmed_control} (d).
A comparative analysis of the pressure coefficient $C_P$ distributions (\autoref{fig:Plano_xy}, top view, and \autoref{fig:Plano_xz}, side view) reveals that all optimised strategies reduce both the spatial extent and intensity of the recirculation bubble compared to the baseline. The most immediate effect of control, visible in every optimised case, is the forward shift of the pressure recovery region at the rear surface, and a partial or near-complete attenuation of the low‑pressure zone characteristic of bluff-body wakes. Notably, the standard GA strongly suppresses the recirculation bubble, resulting in elevated rear-surface pressures and an apparent minimisation of the pressure differential across the body. Nevertheless, this outcome is associated with the creation of concentrated under‑pressure regions, particularly along the vertical edges and near the junction of the slanted and vertical rear surfaces (\autoref{fig:Plano_xz}). These local deficits are indicative of strong, adverse pressure gradients and intense corner separations, suggesting the GA achieves recirculation suppression through energetically costly, non-cooperative actuation in which large jet velocities (notably $U_4$) are directed inwards. The resultant flow closely resembles a geometric boat-tailing, where streamlines align with the slanted surface and recirculation is minimal, effectively mimicking a sharp triangular trailing edge. This severe separation occurs despite the compensatory efforts of actuators 1 and 2. Additionally, actuators 3 and 5 are oriented to reduce under-pressure on the rear surface, although their actuation magnitudes are considerably smaller. As a result, the total drag reduction achieved by GA remains moderate, despite the successful suppression of the recirculation bubble.
By contrast, the direct \HYGO optimum pursues a more nuanced wake reorganisation, displacing but not eliminating the recirculation zone as shown in \autoref{fig:Plano_xy} and \autoref{fig:Plano_xz}. Moderate actuation magnitudes allow the first and second jets to focus on stabilising the boundary layer and delaying its separation, resulting in expanded pressure recovery zones without the energetic penalties of the GA’s solution. The angular distribution of the jets, with a reversed third actuator (compared to GA), produces a controlled deflection of the vortex core downstream, as highlighted by the diminished vorticity concentrations at the vertical edges (\autoref{fig:Plano_yz}). This leads to a recirculation structure of reduced area and strength, as well as a less severe pressure gradient across the wake. Interestingly, the pressure contours and velocity fields suggest that counter-rotating vortices drive separation at the lateral edges of the slanted surface, a configuration that appears to limit the lateral spread of separation and thereby reduces overall drag contributions.

The stepped-\HYGO approach synthesises the advantageous features of both preceding strategies. \autoref{fig:Plano_xy} demonstrates that this configuration partially suppresses the recirculation bubble, similar to GA, while also incorporating the separation control mechanisms observed in \HYGO. This results in minimal recirculation and under-pressure at the rear (though slightly higher than in GA), and confines separation primarily to the vicinity of the slanted surface. Importantly, the stepped solution confines separation predominantly to the near-wake of the slanted surface, avoiding the broad, persistent vortices of the GA strategy and the residual core retained in \HYGO's solution. The optimised actuation parameters (\autoref{tab:optimization_params_furgo}) reflect a compromise: $U_4$ is sufficiently large to exert decisive control over bubble location yet not so dominant as to induce spurious separations, thus facilitating a cooperative action amongst all jets, mainly allowing $U_1$ and $U_2$ to play a more active role in controlling separation. This cooperative mechanism is further corroborated by the streamwise vorticity fields in \autoref{fig:Plano_yz}, which show that in the stepped configuration, both main and secondary vortices are moderated, and their coherence is disrupted, promoting earlier wake recovery and enhanced pressure reattachment. Such disruption is likely aided by the appearance of secondary, oppositely signed vortex structures between the primary pairs, a feature more prominent in both \HYGO solutions.

The transition from brute-force bubble suppression (GA) to judicious wake restructuring (\HYGO and, particularly, \HYGO-Stepped) is thus made explicit in these comparative flow visualisations. While the GA achieves apparent bubble elimination, its approach incurs a penalty in the form of intense, spatially localised separations that limit its net drag reduction potential and produce more intense and long-lived vortices due to the aggressive separation induced by strong actuation (\autoref{fig:Plano_yz}). The direct \HYGO solution, although not eradicating the bubble, uses lower actuation velocities and precision in jet orientation to yield a more uniform, energetically efficient wake. This leads to a smoother wake profile with less lateral momentum exchange, aligning with the observed lower drag values. Finally, the stepped strategy capitalises on both bubble mitigation and wake tailoring, achieving the largest observed drag reduction through a balance of spatially broad separation control and mitigated vortex amplification.

From the standpoint of optimisation, these results offer concrete evidence that the algorithmic decisions encoded by \HYGO and its stepped variant (namely, the orchestration of actuation velocities and orientations, and the sequential refinement of parameter subspaces) result in more optimal cooperative control regimes. The consequential flow structures are physically distinct: improved pressure recovery, limited adverse gradients, and subdued, fragmented vortical patterns. The importance of deliberately staged actuation is thereby affirmed, with clear manifestations in both global cost reduction and flow-field topology. These insights clarify that algorithmic diversity management and staged exploitation do not merely accelerate convergence, but engender physically superior, more robust optimal states in complex fluid–structure control scenarios.

%----------------------------------------------------------
%----------------------- Conclusions ----------------------
%----------------------------------------------------------
\section{Conclusions}\label{s:conclusions}
This work presents the Hybrid Genetic Optimisation framework, \HYGO, as a robust, extensible solution for high-dimensional and complex global unconstrained optimisation. By hybridising evolutionary algorithms with local refinement, specifically, a degeneracy-proof DSM, \HYGO bridges global exploration with accelerated, reliable local exploitation within a unified, modular framework.

A cornerstone of the methodology is the custom degeneracy-proof extension to DSM. This innovation dynamically detects and corrects simplex degeneracies, ensuring that, even in high-dimensional and ill-conditioned search spaces, the simplex retains full rank and is able to efficiently navigate towards optima rather than becoming trapped in lower-dimensional subspaces. This addresses a common and critical failure mode of conventional simplex approaches, greatly improving robustness and convergence, especially as dimensionality increases.

The proposed framework supports both binary-encoded GA and functional (LGP) representations. \HYGO  is employed on a suite of analytical benchmark functions, ranging up to 25 dimensions and encompassing both multimodal and ill-posed topologies, and demonstrates consistent advantages in both convergence speed and solution quality versus classical GAs. These results were statistically robust across 50-fold repetitions and in direct comparison against state-of-the-art hybrid evolutionary algorithms, including CMA-ES, DE, L-SHADE, and PSO, in high-dimensional settings. \HYGO has also shown remarkable robustness and efficiency across any topology, since its performance remains competitive across all analytical functions compared with L-SHADE, which excels when its history-based adaptation and linear population reduction align with the function's characteristics, but struggles in highly non-convex scenarios. \HYGO's exploitation component (DSM) imparts clear efficiency gains, particularly when the search landscape exhibits strong non-separability and premature convergence. However, the test suite contains inherently smooth optimisation strategies, for a more complete characterisation of \HYGO, comparison with other metaheuristics that excel in non-smooth landscapes.

For functional learning, \HYGO with LGP-DSM was demonstrated on the stabilisation of the nonlinear, time-dependent Landau oscillator, a canonical control benchmark. Here, the hybrid method not only accelerated discovery and refinement of stabilising control laws, but systematically yielded more efficient (sparser, energy-minimising) solutions, outperforming pure LGP in both speed and achievable cost. This result highlights the utility of the LGP-DSM hybrid for automated design of interpretable control policies in nonlinear dynamical systems.

The framework’s efficacy was further established on an engineering task: simulation-based active flow control for drag reduction of an Ahmed body. In this computationally expensive, highly nonlinear, and ten-dimensional optimisation problem, \HYGO achieved robust and physically interpretable actuation strategies, realising over 20\% mean drag reduction. The stepped-\HYGO approach, optimising actuation velocities before full vector expansion to include angles with tailored simplex initialisation, proved critical for escaping local traps and achieving global performance. In direct, fair comparison (identical simulation setup and evaluation budget), \HYGO decisively outperformed the EGM and classical GAs on this benchmark, validating its real-world competitiveness and broad applicability.

Overall, \HYGO offers a computationally efficient, easily generalisable approach to high-dimensional global optimisation. The framework’s degeneracy-proof local search, staged hybridisation protocols, and strong empirical validation on both canonical and engineering benchmarks position it as a valuable tool for a wide array of scientific and computational mechanics applications.
As a comprehensive framework for the population-based hybridisation of parametric and functional optimisation, \HYGO offers several promising avenues for generalisation and future development:
\begin{itemize}
    \item \textbf{Automated Hyperparameter Tuning:} Advanced strategies, such as Explorative Landscape Analysis \citep{Mersmann2011gecco} and Persistent Data Topology \citep{WangTY2023book}, could be integrated to extract intrinsic landscape features directly from optimisation data. Identifying key characteristics, such as convexity, multimodality, and the persistence of optima, would allow for the dynamic fine-tuning of \HYGO's hyperparameters throughout the optimisation process.
    \item \textbf{Adaptive Hybridisation Schemes:} Implementing alternate hybridisation methods with varying degrees of exploration and exploitation could further optimise the search process based on problem complexity (e.g., dimensionality and constraint density). For instance, the recently proposed active subspace method \citep{Romor2024jsc} could be employed to reduce the effective dimensionality of the parameter space.  The scheduling of these algorithms could be intelligently driven by real-time extracted landscape features.
    \item \textbf{Dimensionality Reduction and Staged Learning:} For high-dimensional problems, focusing exploration on a reduced parameter subset can significantly accelerate the learning process. This is exemplified by the staged learning approach discussed in \autoref{s:furgo}. Future iterations could utilise sensitivity or uncertainty analysis to prioritise specific parameters, ensuring computational resources are allocated to the most influential variables.
\end{itemize}
While many of these concepts have been successfully implemented or combined for parametric optimisation, their application to functional optimisation remains a critical frontier. Advancing these methods within the \HYGO framework will enable broader applicability to the emerging automation and complex design challenges prevalent in modern computational engineering.

%%%%%%%%%%%%%%%%%%%%%%%%%%%%%%%%%%%%%%%%%%%%%%%%%%%%%%%%%%%%%%
%%%%%%%%%%%%%%%%%%%%%%% Acknowledgments %%%%%%%%%%%%%%%%%%%%%%
%%%%%%%%%%%%%%%%%%%%%%%%%%%%%%%%%%%%%%%%%%%%%%%%%%%%%%%%%%%%%%
\section*{Declaration of competing interest}
The authors declare that they have no known competing financial interests or personal relationships that could have appeared to influence the work reported in this paper.

\section*{Code and Data availability}
The \HYGO framework is openly available to the research and engineering community under the MIT license, including source code, documentation, and ready-to-use examples. \HYGO can be installed directly from the Python Package Index (PyPI) at \hyperlink{https://pypi.org/project/HYGO/}{pypi.org/project/HYGO/} by executing \texttt{pip install HYGO}. In addition, the full development repository is hosted on GitHub at \hyperlink{https://github.com/ipatazas/HYGO}{github.com/ipatazas/HYGO}, including a suite of example scripts illustrating the usage of \HYGO for both parametric benchmark functions and control law optimisation, including the stabilisation of the Landau oscillator. 

\section*{Acknowledgments}
This activity is part of the project ACCREDITATION (Grant No TED2021-131453B-I00), funded by MCIN/AEI/ 10.13039/501100011033 and by the ``European Union NextGenerationEU/PRTR''.
R.C. acknowledge funding by the Madrid Government (Comunidad de Madrid) under the line ``Incentive for Research of Young Doctors'' of the Pluriannual Agreement with the University Carlos III of Madrid (SOLMETAI-CM-UC3M), within the framework of the 6th Regional Plan for Scientific Research and Technological Innovation (VI PRICIT). The authors thank Prof. Discetti, A. Solera-Rico and V. Duro for their useful comments and suggestions.

\section*{Declaration of generative AI and AI-assisted technologies in the writing process}
During the preparation of this work, the authors used ChatGPT (OpenAI) and Grammarly to check grammar, enhance readability, and improve text clarity. After using these tools, the authors reviewed and edited the content as needed and take full responsibility for the content of the publication.

%%%%%%%%%%%%%%%%%%%%%%%%%%%%%%%%%%%%%%%%%%%%%%%%%%%%%%%%%%%%%%
%%%%%%%%%%%%%%%%%%%%%%%% Bibliography %%%%%%%%%%%%%%%%%%%%%%%%
%%%%%%%%%%%%%%%%%%%%%%%%%%%%%%%%%%%%%%%%%%%%%%%%%%%%%%%%%%%%%%
\bibliographystyle{elsarticle-num-names} 
\bibliography{bib.bib}

%%%%%%%%%%%%%%%%%%%%%%%%%%%%%%%%%%%%%%%%%%%%%%%%%%%%%%%%%%%%%%
%%%%%%%%%%%%%%%%%%%%%%%% Appendix %%%%%%%%%%%%%%%%%%%%%%%%%%%%
%%%%%%%%%%%%%%%%%%%%%%%%%%%%%%%%%%%%%%%%%%%%%%%%%%%%%%%%%%%%%%
\clearpage
\appendix

\section{Optimisation Comparison}\label{anex:Comparison}

\begin{table*}[t]
    \centering
    \caption{Median of the minimal cost $J_{min}$ and required evaluations for convergence in the selected functions for low and high-dimensional analysis. The best results are highlighted in bold.}
    \label{tab:comparison_convergence}
    \resizebox{\textwidth}{!}{%
    \begin{tabular}{|c||c|c|c|c|c|c||c|c|c|c|c|c||}
    \hline
    \multirow{2}{*}{Name} & \multicolumn{6}{c||}{$J_{min}$} & \multicolumn{6}{c||}{Evaluations} \\ \cline{2-13} 
                           & \HYGO           & GA          & CMA-ES           & DE          & PSO & LSHADE           & \HYGO             & GA           & CMA-ES           & DE          & PSO & LSHADE         \\ \hline
Rosenbrock2D & \textbf{0.00e+00} & 6.90e-02 & 3.05e-10 & 7.57e-06 & 3.11e-05 & 1.37e-03 & 2593.0 & 2648.5 & 3100.0 & \textbf{2322.0} & 5000.0 & 5000.0 \\ \hline
Rosenbrock25D & 1.58e+02 & 9.29e+03 & 1.41e+02 & 1.21e+03 & 1.93e+04 & \textbf{2.31e+01} & \textbf{4914.0} & 4951.0 & 5100.0 & 5000.0 & 5000.0 & 5000.0 \\ \hline
Rastrigin2D & \textbf{2.42e-06} & \textbf{2.42e-06} & \textbf{2.42e-06} & 6.52e-04 & 4.84e-06 & 2.51e-03 & \textbf{1266.5} & 1283.5 & 3200.0 & 2019.0 & 5000.0 & 5000.0 \\ \hline
Rastrigin25D & \textbf{0.00e+00} & 6.21e+01 & 1.66e+02 & 1.92e+02 & 1.60e+02 & 5.58e+01 & 5000.0 & \textbf{4788.0} & 5000.0 & 5000.0 & 5000.0 & 5000.0 \\ \hline
NSphere2D & \textbf{0.00e+00} & \textbf{0.00e+00} & 6.28e-18 & 1.14e-07 & 9.01e-04 & 5.09e-07 & \textbf{81.0} & 1261.0 & 1300.0 & 2221.0 & 5000.0 & 5000.0 \\ \hline
NSphere25D & 1.79e-02 & 1.18e-01 & 2.75e-04 & 1.45e+00 & 1.07e+01 & \textbf{3.20e-05} & \textbf{4910.0} & 4951.0 & 5100.0 & 5000.0 & 5000.0 & 5000.0 \\ \hline\hline

    \textbf{Total Wins}    & \textbf{4}            & 2   & 1 & 0 & 0 & 2            & \textbf{4}   & 1 & 0 & 1 & 0 & 0  \\ \hline
    \end{tabular}}
\end{table*}

This appendix presents a detailed comparative analysis of the proposed Hybrid Genetic Optimisation (\HYGO) algorithm against four established metaheuristics: the standard Genetic Algorithm (GA), Covariance Matrix Adaptation Evolution Strategy (CMA-ES), Differential Evolution (DE), Particle Swarm Optimisation (PSO), and L-SHADE. The comparison utilises the three widely accepted benchmark functions (Rosenbrock, Rastrigin, and Sphere) selected for a detailed analysis in \autoref{ss:LGA_analytic_high_d} in both two-dimensional (2D) and twenty-five-dimensional (25D) spaces, with performance assessed over 50 independent runs for each scenario.

\autoref{tab:comparison_convergence} summarises the median of the minimal objective function cost, $J_{min}$, achieved by each algorithm. \HYGO demonstrates superior solution quality, securing the best performance in four out of six benchmark scenarios. Specifically, \HYGO attains the true minimum (or near machine precision) for the Sphere 2D and Rastrigin 25D functions, and achieves the lowest median $J_{min}$ for the highly challenging Rosenbrock 25D function ($1.58 \times 10^2$). L-SHADE is the strongest competitor, outperforming \HYGO in the Rosenbrock and Sphere high-dimensional scenarios due to its history-based parameter adaptation that proves more effective than CMA-ES' memory path exploitation, likely due to its simplicity. However, the poor performance of L-SHADE in high-exploration scenarios demonstrates its targeted nature, whereas \HYGO's overall good performance displays its robustness in any topological scenario.

The right side of \autoref{tab:comparison_convergence} reports the median of the number of function evaluations required for the algorithms to converge or to reach the maximum iteration limit. \HYGO exhibits high efficiency in the low-dimensional multimodal test, converging on the Rastrigin 2D problem in only $1266.5$ evaluations, confirming the efficacy of its hybridisation for quickly refining solutions. \HYGO also achieves the fastest convergence on the Sphere 2D problem ($81$ evaluations), demonstrating its efficiency both in exploration and exploitation-dominated problems, while DE is the most efficient for Rosenbrock 2D ($2322$ evaluations), likely due to the vanishing gradients in the central plateau of the function, which hinders the geometrical approach of DSM. For the computationally demanding 25D problems, all population-based algorithms typically require the maximum allowed number of evaluations ($5000$). \HYGO remains competitive in this regime, requiring $4914$ evaluations for the Rosenbrock 25D problem, demonstrating its ability to sustain an effective search over long execution times, and $4910$ in the Sphere function, proving the efficacy of the DSM enrichment.

These comparative results, corroborated by the convergence dynamics illustrated in \autoref{fig:AnalyticalResults2D25D}, demonstrate the significant capabilities of \HYGO across diverse optimisation topologies. The algorithm consistently outperforms the classic GA optimiser and exhibits strong competitive advantages over several modern, state-of-the-art evolution-based strategies, including DE, PSO, L-SHADE, and CMA-ES, validating the efficacy of the proposed hybridisation approach.

\section{Sensitivity analysis}\label{anex:sensitivity_analysis}

\begin{figure*}[t]
    \centering
    \includegraphics[width=0.9\linewidth]{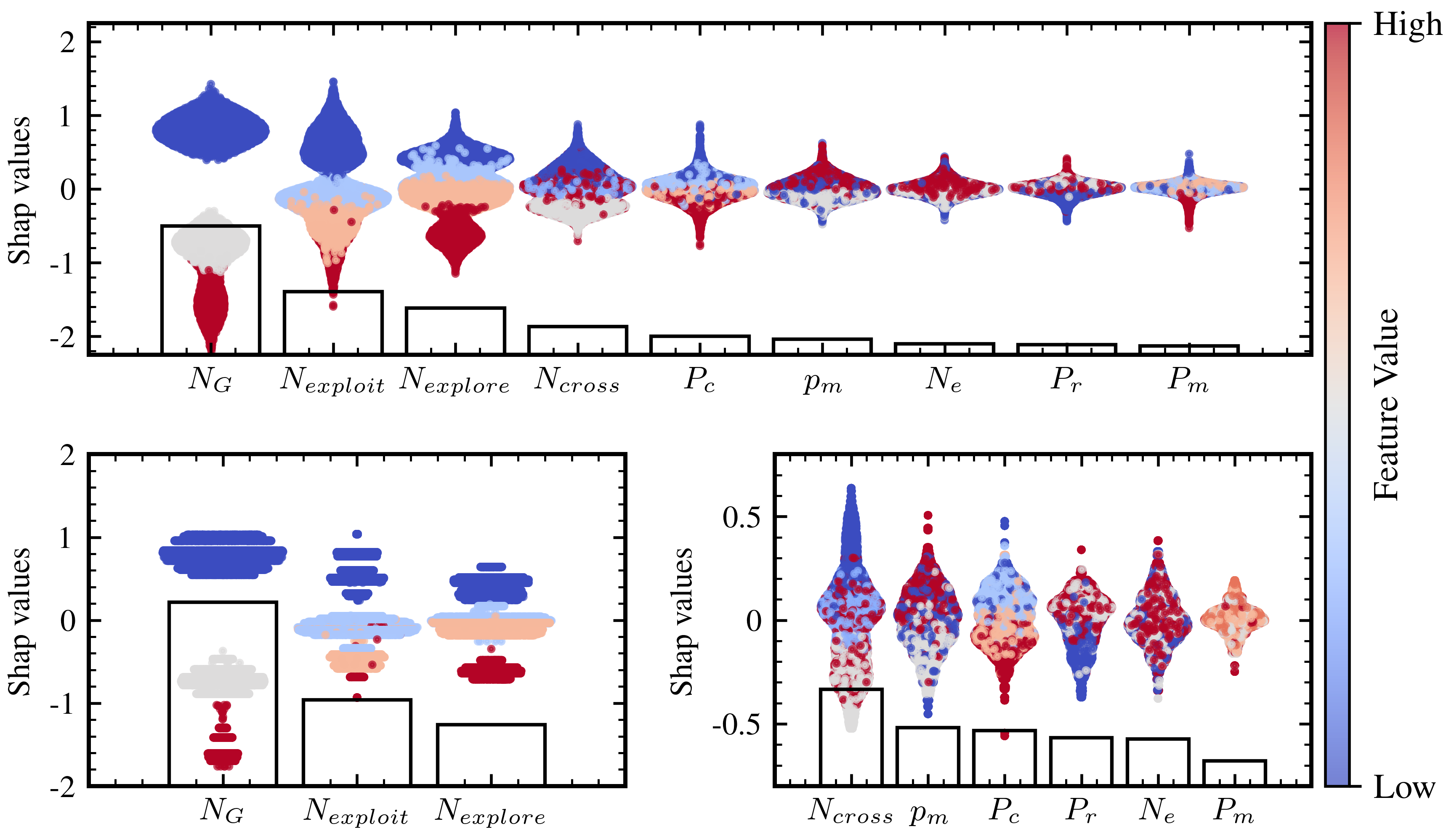}
        \caption{\HYGO Sensitivity analysis. SHAP values for each variable across three distinct analyses. The colour bar indicates the Feature Value (high/low setting) for each parameter. The bars below the violin plots show the mean absolute SHAP value (relative importance).}
    \label{fig:sensitivity_analysis}
\end{figure*}

A comprehensive sensitivity analysis is presented to rigorously determine the influence of the \HYGO optimisation hyperparameters on the final minimal objective function cost ($J_{\min}$). This analysis employs a ``SHapley Additive exPlanations (SHAP)'' methodology \cite{lundberg2017shap}, utilising a  Light Gradient Boosting Machine (LGBM) \cite{ke2017lightgbm} model trained on the converged runs' hyperparameters to predict $J_{\min}$. The analysis was performed across the hyperparameter space defined in \autoref{tab:shap_parameters}, focusing on the Rastrigin function in 7 dimensions for both exploration and exploration capabilities due to its high convexity and intermediate dimensionality.
SHAP assigns an attribution score (SHAP value) to each hyperparameter for every optimisation instance. A \textit{negative} SHAP value with high magnitude indicates that a high parameter setting significantly decreases the target output ($J_{\min}$), thereby expediting convergence. Conversely, a \textit{positive} value indicates the parameter setting increased $J_{\min}$, slowing or hindering performance. 

\begin{table*}
    \centering
    \caption{Hyperparameter space and notation used for the SHAP sensitivity analysis.}
    \label{tab:shap_parameters}
    \resizebox{\textwidth}{!}{%
    \begin{tabular}{llcc}
        \toprule
        \textbf{Parameter Type} & \textbf{Manuscript Parameter} & \textbf{SHAP Notation} & \textbf{Tested Values} \\
        \midrule
        Effort & Number of Generations & $N_G$ & $\{8, 10, 12\}$ \\
        Exploitation & Individuals from DSM & $N_{\text{exploit}}$ & $\{0, 0.05, 0.1\}$ \\
        Exploration & Individuals from Genetic Ops & $N_{\text{explore}}$ & $\{70, 80, 90, 100\}$ \\
        Crossover & Number of Crossover Points & $N_{\text{cross}}$ & $\{1, 2, 3, 5\}$ \\
        Mutation & Mutation Rate & $P_m$ & $\{0.02, 0.05, 0.08\}$ \\
        Elitism & Number of Elitist Individuals & $N_e$ & $\{1, 2, 3\}$ \\
        Genetic Ops Probs & $\text{Pr}, \text{Pc}, \text{Pm}$ (Replication, Crossover, Mutation) & $P_r, P_c, P_m$ & 10 combinations (summing to 1) \\
        \bottomrule
    \end{tabular} }
\end{table*}

The SHAP results are visualised in \autoref{fig:sensitivity_analysis}, which displays three distinct analysis scenarios: a global analysis (top), an analysis focusing on population dynamics (bottom left), and an analysis isolating genetic operators (bottom right). The bars beneath each distribution correspond to the relative weight, or mean absolute SHAP value, of the parameter.
The global analysis reveals that the total search effort and the intrinsic balance between exploration and exploitation are the most critical factors driving the optimisation outcome. The $N_G$ (Number of Generations) is the most influential parameter by a clear margin. The distribution shows that higher values of $N_G$ (Red dots, high Feature Value) are associated with negative SHAP values (low on the $y$-axis), indicating that they drive the cost $J_{\min}$ significantly lower. This is an expected result, as greater evolutionary effort typically yields convergence towards improved marginal solutions due to the increased number of function evaluations.
The $N_{exploit}$ (number of individuals for Local Search/Exploitation) is the second most influential parameter. Higher exploitation individuals (Red dots) are strongly associated with negative SHAP values, suggesting that a higher exploitation effort effectively drives $J_{\min}$ down. Conversely, lower exploitation (Blue dots) generally leads to worse performance, underscoring the necessity of the local search component for solution refinement. The slightly lower impact compared to $N_G$ is likely due to the marginal gains achieved by exploiting an already found local minimum in later generations, where the benefits of global exploration often outweigh further exploitation. Following, the $N_{explore}$ (number of individuals for Exploration based on genetic operations) is the third key driver. High $N_{explore}$ (Red dots) is associated with negative SHAP values, resulting in better performance. This suggests that promoting genetically based exploration is highly beneficial, likely due to the impact a novel individual has when discovering new local minima or when effectively supporting the exploration phase.

Focusing solely on the parameters related to the size of the search space (\autoref{fig:sensitivity_analysis}, bottom left) reinforces the findings of the global analysis. $N_G$ remains dominant, confirming that the total evolutionary effort is the primary determinant of $J_{\min}$. The relative importance and balanced influence of $N_{exploit}$ and $N_{explore}$ also persist in this focused analysis, underscoring the necessity of controlling the exploration-exploitation trade-off.

The internal GA operators (\autoref{fig:sensitivity_analysis}, bottom right) are generally less influential than the effort and hybridisation parameters. The $N_{cross}$ (Number of Crossover Points) is the most important operator setting in this group. While a high number of crossover points does not exhibit a significant performance bias, very few crossover points (e.g., 1) significantly worsen the optimisation output (associated with positive SHAP values). This suggests that more crossover points lead to more efficient mergers of relevant structural information within the generated individuals. On the other hand, the $P_m$ (Mutation Rate) exhibits an interesting nonlinear effect: both very high and very low values have a detrimental effect on $J_{\min}$. This indicates that intermediate mutation rates, where mutated individuals introduce sufficient but not excessive novelty, boost exploration most efficiently, representing an optimal balance. Among the remaining features, the crossover probability ($P_c$) shows a bias favouring higher values to boost convergence, while the replication probability ($P_r$) shows a clear bias favouring minimisation. This aligns with standard evolutionary practice: promoting crossover is essential, while minimising replication is sensible given that the role of preserving highly fit individuals is already fulfilled by elitism.

The SHAP sensitivity analysis provides crucial insights into the \HYGO framework's behaviour. It establishes that the Number of Generations ($N_G$) is the primary driver of the final solution quality due to the increased number of evaluations. Furthermore, the analysis confirms the critical importance of effectively balancing local exploitation ($N_{exploit}$) and global exploration ($N_{explore}$), demonstrating that both higher exploitation and appropriate exploration are necessary to maximise performance. Finally, the analysis guides the selection of internal GA operator settings, highlighting that intermediate mutation rates and sufficient crossover points are optimal. These detailed results validate the design choices of the \HYGO framework and were instrumental in the final parameter selection presented in this paper.
\begin{figure}
    \centering
    \includegraphics[width=0.45\linewidth]{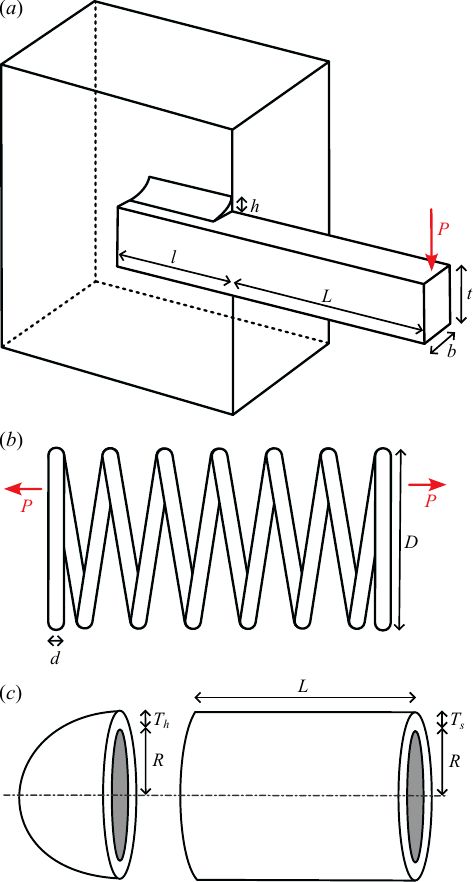}
    \caption{Analytical constrained benchmarks problem sketches, including the Welded beam problem $(a)$, the tension-compression spring problem $(b)$, and the pressure vessel design problem $(c)$. The mathematical formulation of the problems can be found in \citet{kumar2020testconstrained}.}
    \label{fig:CEC}
\end{figure}

\section{Constrained benchmarks: Engineering design problems}
This section evaluates the efficacy of the \HYGO framework when applied to a suite of real-world constrained engineering benchmarks. The selected test cases encompass classic structural optimisation problems, including the design of a pressure vessel, a welded beam, and a tension/compression spring. To accommodate the inherent constraints within the \HYGO architecture, a death penalty method is utilised to transform the constrained problems into an unconstrained formulation, effectively filtering infeasible candidate solutions by assigning them a prohibitive objective cost. The optimisation process is conducted with a total budget of 50,000 function evaluations; the search strategy is configured such that 80\% of the individuals are generated through classical genetic operations to ensure broad global exploration, while the remaining 20\% are refined through local exploitation via the DSM. Detailed mathematical formulations and boundary conditions for these benchmarks are provided in \citet{kumar2020testconstrained}.

\subsection{Welded beam design}
The welded beam design problem, illustrated in \autoref{fig:CEC} (a), aims at minimising the total fabrication cost of the assembly, which incorporates costs associated with material volume and welding labour. This optimisation task is governed by a set of mechanical constraints, including limits on shear stress, bending stress, the buckling load on the bar, and the maximum allowable end deflection. The design is defined by four variables: the weld thickness ($h$), the weld length ($l$), the beam thickness ($t$), and the beam width ($b$). The nonlinear interactions between these variables and the high sensitivity of the stress constraints to geometric changes create a challenging, highly constrained search space.

The performance of \HYGO for the welded beam problem is compared against state-of-the-art specialised metaheuristics in \autoref{tab:cec}. The results indicate that \HYGO successfully converged to a cost of 2.381591, identifying the same basin of attraction as several highly-regarded constrained optimisers, including DSS-DE (2.3810) \citep{zhang2008dssdewelded} and IPSO (2.3810) \citep{he2004ipsowelded}. This achievement is noteworthy given that \HYGO utilises a general-purpose hybridisation of genetic exploration and DSM exploitation, whereas its counterparts employ specialised constraint-handling mechanisms. The identified design variables ($h \approx 0.244$, $l \approx 6.221$, $t \approx 8.294$, $b \approx 0.244$) confirm that the global operators effectively navigated the narrow feasible corridors to reach this local optimum. However, the study also reveals a superior set of algorithms with costs near 1.72, identified by algorithms such as MPA \citep{FARAMARZI2020MPAPressure} and RO \citep{kaveh2012rowelded}. \HYGO’s inability to transition into this superior basin highlights the limitations of the death-penalty management by the genetic operations and geometric-based local-search exploitation. The prohibitive costs assigned to infeasible regions create a rigid boundary that discourages the local search from traversing suboptimal or infeasible paths between basins. In this architecture, discovery of the global 1.72 basin relies almost entirely on the mutation operator. While increasing mutation rates could theoretically facilitate such a jump, it would significantly decrease search efficiency without guaranteeing the discovery of the global edge in such a highly discontinuous landscape.

\subsection{Tension/compression spring design}
The second benchmark evaluates the performance of \HYGO on the tension/compression spring design problem, depicted in \autoref{fig:CEC}(b). The objective is to minimise the total weight of the spring, a task critical for high-performance mechanical systems where mass reduction is paramount. This objective is subject to several physical and operational constraints, including limits on the minimum deflection, maximum shear stress, and the surge frequency. The design space is defined by three continuous variables: the wire diameter ($d$), the mean coil diameter ($D$), and the total number of active coils ($N$). This problem is characterised by a relatively small feasible region, requiring high precision to locate the optimum without violating the governing physical laws.

\HYGO demonstrated exceptional convergence efficiency, identifying a high-quality basin of attraction in only 18,000 evaluations, the lowest computational budget required among all compared metaheuristics. As detailed in \autoref{tab:cec}, the algorithm achieved a final cost of 0.012721 with identified design variables of $d = 0.051429$, $D = 0.349744$, and $N = 11.752381$. While the framework rapidly positioned itself in the immediate vicinity of the global optimum, the precision of the final refinement was influenced by the death penalty constraint-handling strategy. By assigning prohibitive objective costs to even marginal feasibility violations, this method creates a discontinuous landscape at the feasibility interface, which discourages the local search component (DSM) from aggressively exploring the constraint boundaries. Consequently, the DSM routine tends to converge safely within the feasible interior, leaving the final localised fine-tuning dependent on stochastic mutation operators. Despite the slight cost margin relative to specialised algorithms such as MPA \citep{FARAMARZI2020MPAPressure} (0.012665) or T-Cell \citep{aragon2010TcellPressure} (0.012665), these results confirm that \HYGO's hybrid architecture is robust and highly effective at identifying optimal global regions with minimal parameter tuning.

\subsection{Pressure vessel design}
The pressure vessel design problem, illustrated in \autoref{fig:CEC}(c), is a canonical benchmark for mixed-integer optimisation in structural engineering. The primary objective is to minimise the aggregate cost associated with material procurement, welding operations, and shell forming. The design space is defined by four variables: the thickness of the shell, the thickness of the head, the inner radius, and the cylindrical section length. A critical constraint in this problem, governed by ASME boiler code requirements, is that both $T_s$ and $T_h$ must be discrete integer multiples of 0.0625 inches. While some studies in the literature treat this as a continuous optimisation task for simplicity, the discrete nature of the thickness variables significantly complicates the search landscape by introducing discontinuities in the feasibility region.

\HYGO demonstrates a robust capability to navigate the mixed-integer search space associated with the pressure vessel design problem. As evidenced in \autoref{tab:cec}, the algorithm identified a high-quality basin of attraction, yielding a final objective cost of 6188.2547. This performance is within approximately 2\% of the results reported by highly specialised, constraint-tailored metaheuristics such as MPA \citep{FARAMARZI2020MPAPressure} (6059.7144) and DTS-GA \citep{coello2001dtsgaPressure} (6059.9463). Analysis of \HYGO's converged solution ($T_s = 0.9257$, $T_h = 0.4576$, $R = 47.9618$, $L = 115.386$) indicates that the primary deviation from the global optimum occurred in the cylindrical section length ($L$) and the inner radius ($R$). This suggests that while the hybrid architecture is highly effective at identifying core feasible regions, the extreme penalties applied at the constraint boundaries slightly restricted the final fine-tuning of the vessel's geometric proportions. 

Compared to algorithms such as MPA, which utilise random walks to move particles while neglecting infeasible directions, \HYGO relies on chromosomal combinations and mutation operations. While the latter is successful at global exploration, it can struggle to execute the infinitesimal, valid adjustments required to slide along a discontinuous feasibility boundary without triggering a prohibitive penalty. Nevertheless, the proximity of the result to the other solvers highlights \HYGO's efficiency in handling non-convex engineering landscapes characterised by the simultaneous presence of continuous and discrete constraints.

\subsection{Synthesis of Performance and Architectural Implications}

The performance of \HYGO across these benchmarks highlights a fundamental trade-off between algorithmic simplicity and optimisation precision. While specialised metaheuristics often incorporate complex, problem-specific heuristics to navigate feasibility boundaries, \HYGO acts as a robust generalist. The framework's ability to handle the mixed-integer nature of the pressure vessel design without specialised discrete operators demonstrates the inherent flexibility of the hybrid approach. While algorithms like MPA utilise random walks to slide along constraints, \HYGO’s success in these spaces confirms that a well-balanced exploration-exploitation ratio can compensate for the lack of specialised constraint-handling logic.

The synergy between global genetic exploration and local DSM exploitation consistently places \HYGO in the optimal basin of attraction across linear, nonlinear, and mixed-integer landscapes. However, the application of the death-penalty method reveals a specific interaction with \HYGO's geometric operators (due to DSM). Unlike the stochastic random walks in MPA that can effectively \textit{bounce} along a boundary, \HYGO's reliance on crossover and simplex-based local search involves deterministic geometric steps. When these operators encounter the sharp discontinuities created by the death penalty, the resulting cliff in the objective landscape discourages the algorithm from exploring the infeasibility interface. This effectively creates a soft-landing requirement where the search prioritises the safety of the feasible interior. This explains why \HYGO frequently matches the performance of top-tier solvers in identifying the correct region but may exhibit a minor precision gap at the exact constraint interface, where the global minimum often resides.

Despite this, the efficiency of the framework remains a significant advantage. As observed in the tension/compression spring problem, \HYGO achieved near-optimal results with significantly fewer function evaluations than many of its counterparts. This suggests that the hybrid architecture is particularly effective for engineering workflows where finding a high-quality, feasible design quickly is more valuable than the marginal gains of exhaustive fine-tuning at the boundary. Ultimately, these results validate \HYGO as a versatile ``out-of-the-box'' optimiser for engineering challenges. Its ability to remain competitive with highly-advanced, tailored metaheuristics underscores the power of its hybridisation philosophy, offering a reliable solution for complex search spaces with minimal implementation overhead and algorithmic tuning.

\begin{table*}[]
\centering
\caption{Optimum results for the welded beam design, the tension/compression spring, and the pressure vessel design problems, including the found minimum, its associated cost, and the required number of evaluations. \HYGO's performance is compared against state-of-the-art metaheuristics from literature.}
    \label{tab:cec}
    \begin{tabular}{llllllll}
    \toprule
         & Algorithm & $h$ & $l$ & $t$ & $b$ & Cost & Eval, No \\ \midrule
         \multirow{14}{*}{\rotatebox[origin=c]{90}{Welded Beam Design}} & SBM \citep{Akhtar2002SBMPressure} & 0.2407 & 6.4851 & 8.2399 & 0.2497 & 2.4426 & 19,259 \\
         & SA \citep{atiqullah2000sawelded} & 0.2471 &6.1451& 8.2721& 0.2495& 2.4148 & - \\
         & BFOA \citep{mezura2008BFOAPressure} & 0.2057& 3.4711& 9.0367& 0.2057& 2.3868 & 48,000 \\
         & SCA \citep{ray2003scawelded} & 0.2444& 6.2380& 8.2886& 0.2446& 2.3854 & 33,095 \\
         & EA \citep{zhang2009eawelded} & 0.2443& 6.2201& 8.2940& 0.2444& 2.3816 & 28,897 \\
         & T-Cell \citep{aragon2010TcellPressure} & 0.2444& 6.1286& 8.2915& 0.2444& 2.3811 & 320,000 \\
         & FSA \citep{hedar2006fsawelded} & 0.2444& 6.1258& 8.2939& 0.2444& 2.3811 & 56,243 \\
         & IPSO \citep{he2004ipsowelded} & 0.2444& 6.2175& 8.2915& 0.2444& 2.3810 & 30,000 \\
         & DSS-DE \citep{zhang2008dssdewelded} & 0.2444& 6.1275& 8.2915& 0.2444& 2.3810 & - \\
         & HS \citep{LEE2005HSPressure} & 0.2442& 6.2231& 8.2915& 0.2443& 2.3807 & - \\
         & HSA-GA \citep{hwang2006hsagawelded} & 0.2231& 1.5815& 12.8468& 0.2245& 2.2500 & 26,466  \\ 
         %& GSA & 0.182129& 3.856979& 10.0000& 0.202376& 1.879952 & - \\ 
         & RO \citep{kaveh2012rowelded} & 0.203687& 3.528467& 9.004233& 0.207241& 1.735344 & 8000 \\ 
         %& CDE \citep{huang2007cdePressure} & 0.203137& 3.542998& 9.033498& 0.206179& 1.733462 & 240,000 \\ 
         & CPSO \citep{He2007CPSOPressure} & 0.202369& 3.544214& 9.048210& 0.205723& 1.728024 & 200,000 \\ 
         & MPA \citep{FARAMARZI2020MPAPressure} & 0.205728& 3.470509& 9.036624& 0.205730& 1.724853 & 25,000 \\ 
         & \HYGO & 0.244225& 6.221325& 8.294153& 0.244352 & 2.381591 & 50,000 \\ \midrule
         & Algorithm & $d$ & $D$ & $N$ &  & Cost & Eval, No \\ \midrule
         \multirow{12}{*}{\rotatebox[origin=c]{90}{Tension-Compression Spring}} & GA(1)\citep{coello2000ga1tension} & 0.051480& 0.351661& 11.632201 & & 0.012704 & - \\
         & CA \citep{coello2004catension} & 0.050000& 0.317395& 14.031795 & & 0.012721 & 50,000 \\
         %& GSA & 0.050276& 0.323680& 13.525410 & & 0.012702& - \\
         & GA(2) \citep{Bernardino2007hga2Pressure} & 0.051989& 0.363965& 10.890522 & & 0.012681 & 80,000 \\
         & ES \citep{mezura2005estension} & 0.051643& 0.355360& 11.397926 & & 0.012698 & - \\
         & CPSO \citep{He2007CPSOPressure} & 0.051728& 0.357644& 11.244543 & & 0.012674 & 200,000 \\
         & BFOA \citep{mezura2008BFOAPressure} & 0.051825& 0.359935& 11.107103 & & 0.012671 & 48,000 \\
         % & CDE \citep{huang2007cdePressure} & 0.051609& 0.354714& 11.410831 & & 0.012670 & 240,000 \\
         & SCA \citep{ray2003scawelded} & 0.052160& 0.368159& 10.648442 & & 0.012669 & 25,167 \\
         & HGA \citep{bernardino2007hgatension} & 0.051302& 0.347475& 11.852177 & & 0.012668 & 36,000 \\ 
         & PFA \citep{yapici2019pfatension} & 0.051726& 0.357629& 11.235724 & & 0.012665 & - \\ 
         & T-Cell \citep{aragon2010TcellPressure} & 0.051622& 0.355105& 11.384534 & & 0.012665 & 36,000 \\ 
         & MPA \citep{FARAMARZI2020MPAPressure} & 0.051724& 0.357570& 11.239195 & & 0.012665 & 25,000 \\ 
         & \HYGO & 0.051429& 0.349744& 11.752381 &  & 0.012721 & 18,000 \\ \midrule
         & Algorithm & $T_s$ & $T_h$ & $R$ & $L$ & Cost & Eval, No \\ \midrule
         \multirow{13}{*}{\rotatebox[origin=c]{90}{Pressure Vessel}} & HS \citep{LEE2005HSPressure} & 1.125 & 0.625 & 58.2789 & 43.7549 & 7198.433 & - \\
         % & GSA & 1.125 & 0.625 & 55.9887 & 84 0.4542 & 85380.8359 & - \\
         & HGA(2) \citep{Bernardino2007hga2Pressure} & 1.1250 & 0.5625 & 58.1267 & 44.5941 & 6832.583 & 80,000 \\
         & GeneAS \citep{Deb1997geneasPressure} & 0.9375 & 0.5000 & 48.3290 & 112.6790 & 6410.3811 & - \\
         & T-Cell \citep{aragon2010TcellPressure} & 0.8125 & 0.4375 & 42.098429 & 190.787695 & 6390.554 & 80,000 \\
         & SBM \citep{Akhtar2002SBMPressure} & 0.8125 & 0.4375 & 41.9768 & 182.2845 & 6171.000 & 12,630 \\
         & HGA(1) \citep{Bernardino2007hga2Pressure} & 0.8125 & 0.4375 & 42.0492 & 177.2522 & 6065.821 & 80,000 \\
         & CPSO \citep{He2007CPSOPressure} & 0.8125 & 0.4375 & 42.091266 & 176.746500 & 6061.0777 & 200,000 \\
         & BFOA \citep{mezura2008BFOAPressure} & 0.8125 & 0.4375 & 42.096394 & 176.683231 & 6060.460 & 48,000 \\
         & HAIS-GA \citep{coello2004haisgaPressure} & 0.8125 & 0.4375 & 42.0931 & 176.7031 & 6060.367 & 150,000 \\ 
         & DTS-GA \citep{coello2001dtsgaPressure} & 0.8125 & 0.4375 & 42.097398 & 176.654047 & 6059.9463 & 80,000 \\ 
         & ES \citep{mezura2008esPressure} & 0.8125 & 0.4375 & 42.098087 & 176.640518 & 6059.745 & 25,000 \\ 
         % & CDE \citep{huang2007cdePressure} & 0.8125 & 0.4375 & 42.098411 & 176.637690 & 6059.7340 & 250,000 \\ 
         & MPA \citep{FARAMARZI2020MPAPressure} & 0.8125 & 0.4375 & 42.098445 & 176.636607 & 6059.7144 & 25,000 \\ 
         & \HYGO & 0.9257&   0.4576&  47.9618& 115.386 & 6188.2547 & 50,000 \\ \bottomrule
    \end{tabular}
\end{table*}

\end{document}